%% file: iclr2026_conference.tex
\documentclass{article} 
\usepackage{iclr2026_conference,times}

\input{math_commands.tex}

\usepackage{hyperref}
\DeclareRobustCommand{\eqref}[1]{\textup{{Eq. (\ref{#1})}}}
\usepackage{url}
\usepackage[utf8]{inputenc} 
\usepackage{hyperref}       
\usepackage{url}            
\usepackage{booktabs}       
\usepackage{amsfonts}       
\usepackage{nicefrac}       
\usepackage{microtype}      
\usepackage{xcolor}         
\usepackage{latexsym}
\usepackage{amssymb}
\usepackage{caption}
\usepackage{subcaption}
\usepackage{amsmath}
\usepackage{amsthm}
\usepackage{enumitem}
\usepackage{color}
\usepackage{algorithm}
\usepackage{subcaption}
\usepackage{graphicx}
\usepackage{listings}
\usepackage{algorithmic}
\usepackage{multirow}
\usepackage{siunitx} 
\hypersetup{hidelinks,
    colorlinks=true,
    anchorcolor=blue,
    citecolor={olive},
    linkcolor=blue}
    
\newtheorem{theorem}{Theorem}

\usepackage{wrapfig}
\usepackage{siunitx} 
\title{Enhanced DACER Algorithm with High Diffusion Efficiency}

\author{Yinuo Wang\textsuperscript{1*}, Likun Wang\textsuperscript{1*}, Mining Tan\textsuperscript{2*}, Wenjun Zou\textsuperscript{1}, Xujie Song\textsuperscript{1},\\
\textbf{Wenxuan Wang\textsuperscript{1}}, \textbf{Tong Liu\textsuperscript{1}}, \textbf{Guojian Zhan\textsuperscript{1}}, \textbf{Tianze Zhu\textsuperscript{1}}, \textbf{Shiqi Liu\textsuperscript{1}}, \\
\textbf{Zeyu He\textsuperscript{1}}, \textbf{Feihong Zhang\textsuperscript{1}}, \textbf{Jingliang Duan\textsuperscript{1,3}}, \textbf{Shengbo Eben Li\textsuperscript{1}} \\
\\
\textsuperscript{1}School of Vehicle and Mobility \& College of AI, Tsinghua University \\
\textsuperscript{2}School of Artificial Intelligence, University of Chinese Academy of Sciences \\
\textsuperscript{3}School of Mechanical Engineering, University of Science and Technology Beijing \\
}

\iclrfinalcopy 
\begin{document}

\maketitle

\begin{abstract}
Due to their expressive capacity, diffusion models have shown great promise in offline RL and imitation learning. Diffusion Actor-Critic with Entropy Regulator (DACER) extended this capability to online RL by using the reverse diffusion process as a policy approximator, achieving state-of-the-art performance. However, it still suffers from a core trade-off: more diffusion steps ensure high performance but reduce efficiency, while fewer steps degrade performance. This remains a major bottleneck for deploying diffusion policies in real-time online RL. To mitigate this, we propose DACERv2, which leverages a Q-gradient field objective with respect to action as an auxiliary optimization target to guide the denoising process at each diffusion step, thereby introducing intermediate supervisory signals that enhance the efficiency of single-step diffusion. Additionally, we observe that the independence of the Q-gradient field from the diffusion time step is inconsistent with the characteristics of the diffusion process. To address this issue, a temporal weighting mechanism is introduced, allowing the model to effectively eliminate large-scale noise during the early stages and refine its outputs in the later stages. Experimental results on OpenAI Gym benchmarks and multimodal tasks demonstrate that, compared with classical and diffusion-based online RL algorithms, DACERv2 achieves higher performance in most complex control environments with only \textbf{five diffusion steps} and shows greater multimodality.
\end{abstract}

\section{Introduction}
\label{sec: Introduction}

\begin{figure}[ht!]
  \centering
    \includegraphics[width=0.65\textwidth]{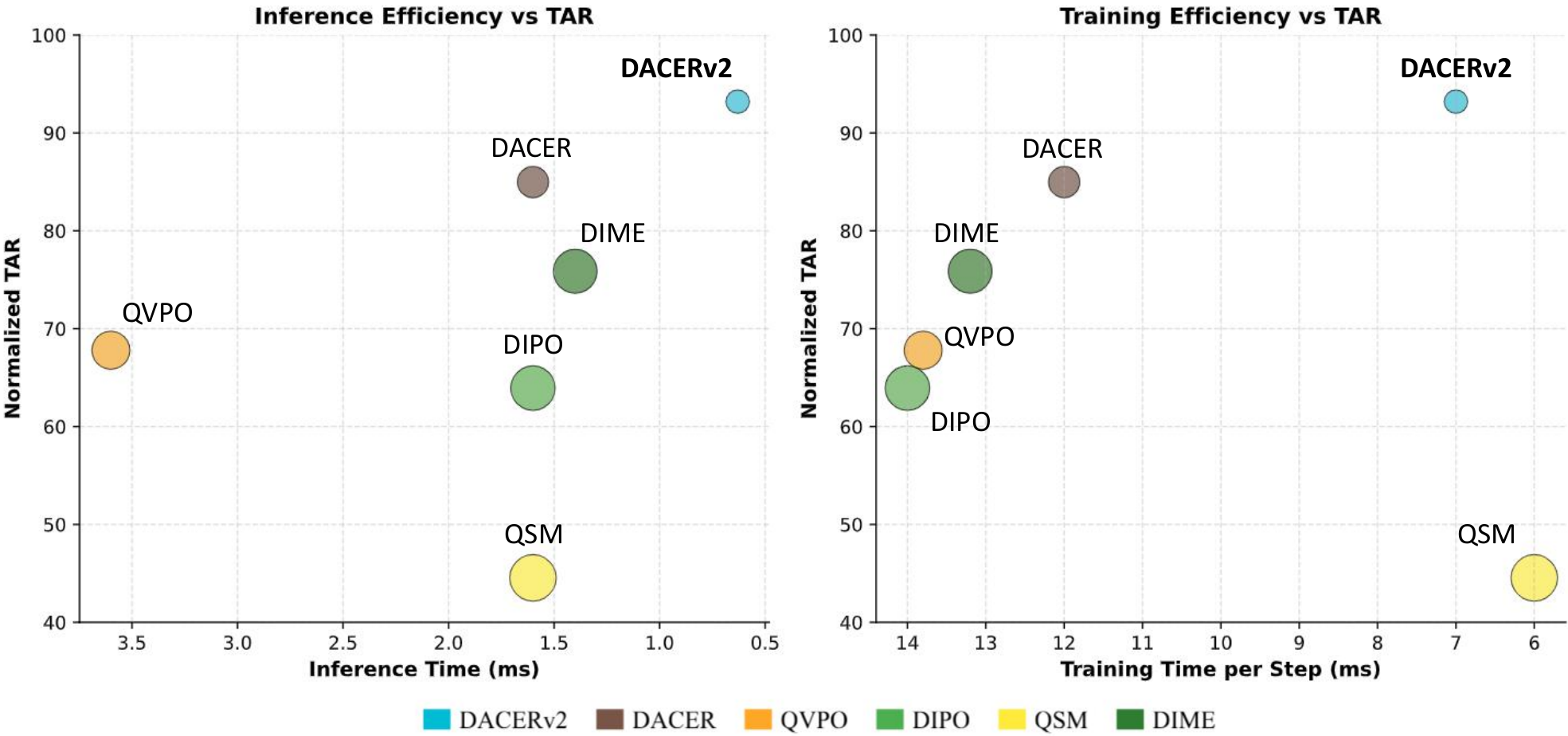} 
  \caption{\textbf{Efficiency and Performance.} The horizontal axis represents the training or inference time (increasing from right to left), while the vertical axis shows the normalized Total Average Return (TAR). The training time is the per-step computational cost on OpenAI Gym tasks, excluding the
time spent on environment interaction. The inference time is measured as the latency required for
the policy network to output an action given a single state as input. DACERv2 achieve outstanding performance.}
\label{fig:comparison_time}
\end{figure}

Energy-based models are well-suited as agent policy functions due to their powerful representational capailities. Learning a policy to approximate the corresponding energy-based target distribution allows for modeling complex and multimodal action patterns without relying on restrictive parametric assumptions, especially in continuous action spaces. This enhanced expressiveness can significantly improve exploration by enabling the agent to discover and leverage diverse behavioral strategies. However, effectively approximating such an expressive soft policy presents notable challenges. A key difficulty lies in how to efficiently and accurately sample from the target distribution. While algorithms such as Soft Actor-Critic (SAC) \citep{haarnoja2018soft} and Distributional Soft Actor-Critic (DSAC) \citep{duan2021distributional,duan2023dsac} aim to approximate the soft-target distribution, they typically represent the policy as a simple Gaussian, enabling analytical entropy computation. This choice is computationally efficient but falls short in modeling complex and multimodal behavior. Meanwhile, due to their strong representational capacity, diffusion models have emerged as a promising policy class for continuous control, commonly referred to as diffusion policies \citep{ren2024DPPO, li2024learning, lu2025makes}. 

Existing methods for training diffusion policies can be broadly categorized into two groups: score-matching and end-to-end policy gradient approaches. In the first group, QVPO \citep{ding2024diffusion} proposes using Q-weighted imitation learning samples to improve policy learning. QSM \citep{psenka2023learning} directly aligns the score functions with the gradients of the learned Q-functions and uses Langevin dynamics for sampling. DIPO \citep{yang2023policy} updates the replay buffer using action gradients and improves the performance of the policy through a diffusion loss. In the second group, DACER \citep{wang2024diffusion} directly backward the gradient through the reverse diffusion process and proposes a Gaussian mixture model (GMM) entropy regulator to balance exploration and exploitation. DIME \citep{celik2025dime} derives an approximate maximum-entropy lower bound, directly integrating the maximum-entropy RL framework with the diffusion policy. However, diffusion policies typically require a large number of diffusion steps to maintain strong performance, which results in low inference efficiency. Although acceleration techniques such as DPM-Solver \citep{lu2022dpm} can reduce the number of diffusion steps, this often comes at the cost of performance degradation. As a result, previous methods struggle to escape the dilemma between performance and time-efficiency.

To tackle this issue, we present DACERv2, a highly efficient diffusion-based RL algorithm that achieves comparable or superior performance with only a few diffusion steps, as shown in Fig.~\ref{fig:comparison_time}. The key contributions of this paper are the following: 1) We propose a Q-gradient field objective as an extra intermediate supervisory signal to enhance the efficiency of single-step diffusion. 2) Since the Q-gradient field is independent of the diffusion time, we propose a temporal weighting mechanism that takes the current diffusion time step as input. This mechanism aligns with the requirements of the diffusion denoising process: higher denoising amplitudes during early stages and lower denoising amplitudes for precise control in later stages. 3) We evaluate the performance of our method on the OpenAI Gym benchmark \citep{brockman2016openai}. Compared with both diffusion-based and classical algorithms like DACER \citep{wang2024diffusion}, QVPO \citep{ding2024diffusion}, DIME \citep{celik2025dime}, QSM \citep{psenka2023learning}, DIPO \citep{yang2023policy}, DSAC \citep{duan2023dsac}, PPO \citep{schulman2017proximal}, and SAC \citep{haarnoja2018soft}, our approach achieved state-of-the-art (SOTA) performance in most complex control tasks. 4) We evaluate the training and inference times of all diffusion-based algorithms under identical hardware configurations using the PyTorch framework. While achieving stronger overall performance, our method reduces training time by \textbf{47.0\%} and \textbf{41.7\%}, and inference time by \textbf{55.0\%} and \textbf{60.6\%}, compared with DIME and DACER, respectively.

\section{Preliminaries}
\label{sec: Preliminary}
\subsection{Reinforcement Learning with Soft Policy}
\label{pre: soft policy}

RL problems are commonly modeled as Markov Decision Processes (MDPs) \citep{sutton2018reinforcement, li2023reinforcement}. An infinite-horizon MDP is defined by a tuple $(\mathcal{S}, \mathcal{A}, P, r, \gamma)$, where $\mathcal{S}$ is the state space and $\mathcal{A}$ is the action space, both assumed bounded and potentially continuous. $P: \mathcal{S} \times \mathcal{A} \mapsto \Delta(\mathcal{S})$ denotes the transition dynamics, specifying the probability distribution $P(\cdot \mid s_t, a_t)$ over next states, with $\Delta(\mathcal{S})$ representing the set of distributions over $\mathcal{S}$. $r: \mathcal{S} \times \mathcal{A} \mapsto \mathbb{R}$ is the reward function, and $\gamma \in [0,1)$ is the discount factor. The behavior of agent  is characterized by a policy $\pi: \mathcal{S} \mapsto A$, which defines the process of selecting action $a$ given the state $s$.

To evaluate the value of taking an action $a$ in a given state $s$ under policy $\pi$, the action-value function $Q^\pi(s, a)$ is introduced, which represents the expected cumulative discounted reward, defined as follows:
\begin{equation}
Q^\pi(s,a) = \mathbb{E}_{\pi} \Big[ \sum_{i=0}^{\infty} \gamma^i r(s_i, a_i) \mid s_0=s, a_0=a \Big].
\end{equation}

A key challenge in online RL is the trade-off between exploration, gathering information for future gains, and exploitation, maximizing returns based on current knowledge. One compelling strategy involves learning a policy that aims to approximate a soft policy \citep{haarnoja2017reinforcement,haarnoja2018soft,ma2025soft,messaoud2024s}. Such target soft policies are typically formulated as a Boltzmann distribution, where the desired policy distribution is proportional to the exponentiated state-action value function:
\begin{equation}
  \pi_{\text{soft}}(a|s) \propto \exp\left(\frac{1}{\alpha} Q(s,a)\right).
\end{equation}

The target of soft policy is to minimize the per-state KL divergence $D_{\mathrm{KL}}\!\left(\pi(\cdot|s)\,\|\,\frac{\exp(Q(s,\cdot)/\alpha)}{Z(s)}\right)$, where $Z(s)$ is the normalization coefficient. This KL-divergence minimization problem is equivalent to maximizing a final objective function that balances value maximization and entropy regularization:
\begin{equation}
J(\pi) = \mathbb{E}_{(s,a) \sim \pi} 
\big[ Q(s,a) \big] + \alpha \cdot \mathcal{H}(\pi(\cdot|s)).
\end{equation}

Diffusion policies are able to model complex policy distributions, but their entropy is analytically intractable. Fortunately, in methods like DACER \citep{wang2024diffusion}, maximizing the $Q$-value objective under specific entropy regularization likewise produces a Boltzmann policy. See Theorem \ref{the: soft policy} for further theoretical details.

\subsection{Diffusion Models as Expressive Policy}

Diffusion models \citep{ho2020denoising, song2020score, wang2024diffusion} conceptualize data generation as a stochastic process where data samples are iteratively reconstructed via a parameterized reverse-time stochastic differential equation (SDE). Although both forward and reverse diffusion processes are theoretically integral to these models, recent work \citep{chen2024deconstructing} highlights that their expressive power primarily stems from the reverse-time denoising dynamics, rather than the forward-time noising process. Accordingly, our analysis and modeling efforts concentrate on the reverse diffusion process.

Formally, the continuous reverse-time SDE that governs this process is defined as follows:
\begin{equation}
d\boldsymbol{x} = \left[ f(\boldsymbol{x}, t) - g(t)^2 \nabla_{\boldsymbol{x}} \log p_t(\boldsymbol{x}) \right] dt + g(t) \, d\omega(t),
\end{equation}
where $f(\boldsymbol{x}, t)$ represents the drift term, $g(t)$ denotes the time-dependent diffusion coefficient, $\nabla_{\boldsymbol{x}} \log p_t(\boldsymbol{x})$ is the score function, and $d\omega(t)$ is the standard Wiener process. The term $\nabla_{\boldsymbol{x}} \log p_t(\boldsymbol{x})$, also known as the score function, plays a central role in guiding the reverse diffusion dynamics. It is important to note that this equation represents the general form of the reverse-time SDE; the specific construction of terms such as $f(\boldsymbol{x}, t)$ and $g(t)$ can vary across different diffusion model algorithms.

Therefore, the reverse-time SDE of diffusion policy can be expressed as:
\begin{equation}
\label{eq: sde policy}
d\boldsymbol{a}_t = \left[ f(\boldsymbol{a}_t, t) - g(t)^2 S_{\theta}(\boldsymbol{s}, \boldsymbol{a}_t, t)  \right] dt + g(t) \, d\omega(t),
\end{equation}
where $S_{\theta}(\boldsymbol{s}, \boldsymbol{a}_t, t)$ is a neural network designed to approximate the gradient $\nabla_{\boldsymbol{a}_t} \log p_t(\boldsymbol{a}_t|\boldsymbol{s})$. Actions can be sampled from the diffusion policy $\pi_{\theta}(\boldsymbol{a}_0|\boldsymbol{s})$ by solving the following integral:
\begin{equation}
\label{eq: diffusion policy sde}
\boldsymbol{a}_0
  = \boldsymbol{a}_T
  + \int_{0}^{T} \!\left[
      f\!\left(\boldsymbol{a}_{\tau},\tau\right)
      - g(\tau)^{2}\,
        S_{\theta}\!\left(\boldsymbol{s}, \mathbf{a}_{\tau},\tau\right)
    \right] d\tau
  + \int_{0}^{T} g(\tau)\, d\omega(\tau),
\end{equation}
where $\boldsymbol{a}_T$ follows a standard normal distribution $\mathcal{N}(0, \boldsymbol{I})$.

\subsection{Langevin Dynamics}
Langevin dynamics represents a powerful computational framework for simulating particle motion under the joint influence of deterministic forces and stochastic fluctuations. When coupled with stochastic gradient descent, this approach gives rise to stochastic gradient Langevin dynamics (SGLD) \citep{welling2011bayesian} - an efficient sampling algorithm that leverages log-probability gradients $\nabla_{\boldsymbol{x}}\log p(\boldsymbol{x})$ to draw samples from probability distributions $p(x)$ through an iterative Markov chain process:
\begin{equation}
\label{eq: Langevin}
\boldsymbol{x}_{t-1}=\boldsymbol{x}_{t}+\frac{\delta_t}{2}\nabla_{\boldsymbol{x}_t}\log p(\boldsymbol{x}_{t})+\sqrt\delta_t\boldsymbol{\epsilon},
\end{equation}
where $\boldsymbol{\epsilon}\sim\mathcal{N}(\boldsymbol{0},\boldsymbol{I})$, $\delta_t$ is the step size. When $t$ range from infinity to one,  $\delta_t \to0$, $x_0$ equals to the true probability density $p(x)$.

\section{Method}
\label{sec: method}
In this section, we explain how our method approximates the target policy distribution with fewer diffusion steps. First, we show that $\nabla_{\boldsymbol{a}_t} Q(\boldsymbol{s}, \boldsymbol{a}_t)$, derived from Langevin dynamics, can be incorporated into the unified SDE-based framework for action generation, thereby improving the efficiency of single-step diffusion. However, when this extra objective function is incorporated, the diffusion policy only exhibits suboptimal performance. This limitation arises because $\nabla_{\boldsymbol{a}_t} Q(\boldsymbol{s}, \boldsymbol{a}_t)$ remains independent of the diffusion step, whereas the score function is not. Therefore, we introduce a time-weighted mechanism to better align with the requirements of the diffusion denoising process. Lastly, we propose a practical algorithm for optimizing diffusion models.

\subsection{Q-Gradient Field Guided Denoising}
\label{sec:q-gradient field}
Using the only reverse process, the objective function of DACER is to maximize the Q-value, representing an end-to-end optimization approach without direct supervision in the intermediate diffusion steps. However, within this optimization scheme, the guidance signals at intermediate steps are implicit, as they are derived solely from the final Q-value through back-propagation, which in turn necessitates more diffusion steps to produce higher-quality control actions. To address this issue, we propose the Q-gradient field function as an extra training loss to enhance the efficiency of single-step diffusion. At the end of Section \ref{pre: soft policy}, we explain why, when the global policy entropy is fixed, the optimal policy for maximizing the Q-value theoretically follows a Boltzmann distribution. Importantly, this conclusion holds for policy families of arbitrary forms and naturally suits the SDE-based policy families.

From another perspective, Langevin dynamics can be regarded as a special form of an SDE-based policy, providing an efficient method for sampling actions from Boltzmann distributions \citep{hinton2002training}:
\begin{equation}
\label{eq: boltzmann}
\pi(\boldsymbol{a} | \boldsymbol{s}) = \frac{e^{\frac{1}{\alpha} Q(\boldsymbol{s}, \boldsymbol{a})}}{Z(\boldsymbol{s})},
\end{equation}
where $\alpha > 0$ is the temperature coefficient, $Q(\boldsymbol{s}, \boldsymbol{a})$ is the state action value function and $Z(\boldsymbol{s})$ is the partition function that normalizes the distribution. The formula derived by taking the partial derivative of both sides of \eqref{eq: boltzmann} with respect to $\boldsymbol{a}$ can be expressed as
\begin{equation}
\label{eq: boltzmann qiudao}
\nabla_{\boldsymbol{a}}\log \pi (\boldsymbol{a}|\boldsymbol{s}) = \frac{1}{\alpha} \nabla_{\boldsymbol{a}}Q(\boldsymbol{s}, \boldsymbol{a}).
\end{equation}
Substituting \eqref{eq: boltzmann qiudao} into \eqref{eq: Langevin}, we can obtain the sampling process for $\pi (\boldsymbol{a}|\boldsymbol{s})$:
\begin{equation}
\label{eq: Langevin RL}
\boldsymbol{a}_{t-1}=\boldsymbol{a}_{t}+\frac{\delta_t}{2\alpha} \nabla_{\boldsymbol{a}}Q(\boldsymbol{s}, \boldsymbol{a})+\sqrt\delta_t\boldsymbol{\epsilon}.
\end{equation}

In summary, Langevin dynamics can be regarded as a particular solution within the family of SDE-based policies. This connection motivates the use of $\nabla_{\boldsymbol{a}} Q(\boldsymbol{s}, \boldsymbol{a})$ as an extra learning objective to guide the training of SDE-based policies, thereby introducing additional supervision signals into the intermediate diffusion step. Consequently, the efficiency of single-step diffusion can be improved, enabling comparable or even superior performance to previous algorithms with fewer diffusion steps. 

Moreover, in highly unstable environments that exhibit extreme sensitivity to minor action perturbations, the Q-gradient estimation can become volatile, potentially hindering the algorithm's convergence to the optimal policy \citep{ding2024diffusion, ma2025soft}. When the diffusion process is restricted to only a few steps, a policy trained solely on the Q-gradient often struggles to converge. For these reasons, we adopt it only as an auxiliary guidance in policy training.

\subsection{Time-weighted mechanism}
\label{sec:time-weighted}
In the previous subsection, we propose incorporating the gradient term $\nabla_{\boldsymbol{a}_t} Q(\boldsymbol{s}, \boldsymbol{a}_t)$ as an auxiliary objective when training the SDE-based policy. However, experimental results show that directly employing this objective yields suboptimal performance, as shown in Fig. \ref{fig: DACERv2 Ablation}(b). We attribute this to the Q-gradient being independent of the diffusion time step, whereas the score function is not. Such time invariance fails to satisfy the varying denoising requirements across the diffusion process. Specifically, in the later stages of diffusion process, the denoising intensity should naturally decrease as the action distribution approaches the optimal policy.

To address this issue, we introduce a time-weighted mechanism that modulates the influence of Q-gradient guidance based on the diffusion time step, allowing for more precise control over the denoising process. Inspired by the design approach for the step size $\delta_t$ in \eqref{eq: Langevin}, we can similarly design our time-weighted mechanism using the commonly employed exponential decay function \citep{welling2011bayesian, teh2016consistency}:
\begin{equation}
\label{eq: time-weighted}
w(t)=\exp(c \cdot t+d),
\end{equation}
where $t$ denotes the current diffusion step. The hyperparameters $c$ and $d$ are chosen inspired by the variance-preserving beta schedule in DDPM \citep{ho2020denoising} and depend only on the number of diffusion steps. The specific setting is presented in Appendix \ref{app:hyper}.

Furthermore, to improve the stability of the training process, we normalize $\nabla_{\boldsymbol{a}_t} Q(\boldsymbol{s}, \boldsymbol{a}_t)$ by its norm:
\begin{equation}
\label{eq: normalized Q-gradient}
\nabla_{\boldsymbol{a}_t}Q_\text{norm}(\boldsymbol{s}, \boldsymbol{a}_t) = \frac{\nabla_{\boldsymbol{a}_t}Q(\boldsymbol{s}, \boldsymbol{a}_t)}{||\nabla_{\boldsymbol{a}_t}Q(\boldsymbol{s}, \boldsymbol{a}_t)|| + \epsilon},
\end{equation}
where $\epsilon$ is a small constant to prevent division by zero. 

Ultimately, we construct the Q-gradient field objective function to facilitate the training of the diffusion policy, where $\pi_{\theta}(\boldsymbol{a}_t|\boldsymbol{s})$ denotes the action generated using the diffusion policy as defined in \eqref{eq: diffusion policy sde}:

\begin{equation}
\label{eq: q-gradient field}
\mathcal{L}_g(\theta)= \min_\theta\underset{\substack{\boldsymbol{s} \sim \mathcal{B} \\ t \sim \text{U}(1, T) \\ \boldsymbol{a}_t \sim \pi_{\theta}(\boldsymbol{a}_t | \boldsymbol{s})}}{\mathbb{E}}\left[\left\|w(t)\nabla_{\boldsymbol{a}_{t}} Q_\text{norm}(\boldsymbol{s}, \boldsymbol{a}_t) - S_\theta(\boldsymbol{s}, \boldsymbol{a}_t,t)\right\|_2^2\right],
\end{equation}
where U means uniform distribution, $t$ is the current diffusion step, $\mathcal{B}$ represents the replay buffer, and $\theta$ is the network parameter of the diffusion policy. The subscript $g$ represents the objective function related to the Q-gradient.

\subsection{DACERv2: A High Efficiency Diffusion RL Algorithm}
\label{sec:DACERv2}
To obtain a practical algorithm, we use a parameterized function approximation for the Q-function and the diffusion policy. In the critic component, we adopt the double Q-learning strategy \citep{fujimoto2018addressing} to alleviate overestimation bias. Specifically, we maintain two independent Q-function estimators, denoted as $Q_{\phi_1}(\boldsymbol{s}, \boldsymbol{a})$ and $Q_{\phi_2}(\boldsymbol{s}, \boldsymbol{a})$, which are trained to approximate the true action-value function. To enhance training stability, we introduce two corresponding target networks, $Q_{\bar{\phi}_1}(\boldsymbol{s}, \boldsymbol{a})$ and $Q_{\bar{\phi}_2}(\boldsymbol{s}, \boldsymbol{a})$, which are updated softly from the main networks following the technique in \citep{van2016deep}. 

The Q-networks are optimized by minimizing the Bellman error. For each network $Q_{\phi_i}(\boldsymbol{s}, \boldsymbol{a})$, the loss $J_Q(\phi_i)$ is defined as:
\begin{equation}
\label{eq:critic_loss}
J_Q(\phi_i) \!=\! \underset{\substack{(\boldsymbol{s}, \boldsymbol{a}, \boldsymbol{r}, \boldsymbol{s}') \sim \mathcal{B} \\ \boldsymbol{a}' \sim \pi_{\theta}(\boldsymbol{a}_0 | \boldsymbol{s})}}{\mathbb{E}} \left[ \!\left(\! \left(r(\boldsymbol{s}, \boldsymbol{a}) \!+\! \gamma \min_{i=1,2} Q_{\bar{\phi}_i}(\boldsymbol{s}', \boldsymbol{a}') \!\right)-\! Q_{\phi_i}(\boldsymbol{s}, \boldsymbol{a}) \!\right)^2 \right], \\
\end{equation}
where $\gamma$ is discount factor, the target is computed as the smaller of the two target Q-values, $Q_{\bar{\phi}_1}(\boldsymbol{s}', \boldsymbol{a}')$ and $Q_{\bar{\phi}_2}(\boldsymbol{s}', \boldsymbol{a}')$, to prevent over-optimistic estimates. Furthermore, we incorporate the distributional value estimation framework from DSAC \citep{duan2023dsac} to further mitigate overestimation issues.

In the actor component, we follow the objective function of maximizing the Q value and combine it with the auxiliary learning objective based on the Q-gradient field proposed in this paper. The final policy-learning objective is a linear combination: 
\label{eq: final policy-learning objective}
\begin{equation}
\begin{aligned}
\pi = \arg\min_{\pi_\theta} \mathcal{L_\pi}(\theta) = \mathcal{L}_q(\theta) + \eta \cdot \mathcal{L}_g(\theta), \\
\text{s.t.} \quad \mathbb{E}_{s \sim p(s)}[H(\pi^*(\cdot|s))] = \mathcal{H}^{\text{target}},
\end{aligned}
\end{equation}
where $\eta$ is a hyperparameter, $    \mathcal{L}_q(\theta) =
    \mathbb{E}_{\boldsymbol{s}\sim \mathcal{B}, \boldsymbol{a}_0\sim \pi_\theta(\cdot | \boldsymbol{s})}
    \left[-Q_\phi(\boldsymbol{s}, \boldsymbol{a}_0)\right],$ $p(s)$ is a distribution over states. $\pi^*$ is the Boltzmann-optimal policy under the global entropy $\mathcal{H}^{\text{target}}$ . We adopt the entropy regularization method from the original DACER algorithm to control the global policy entropy.

\section{Experimental Results}
\label{sec:exp}
Multimodality is a key metric for evaluating diffusion-based algorithms. Therefore, we first validate DACERv2 with respect to this metric in the “Multi-goal” environment \citep{haarnoja2017reinforcement}, as illustrated in Fig. \ref{fig:multigoal}. We then conducted experiments on eight tasks in OpenAI Gym MuJoCo \citep{2012MuJoCo}. These environments represent challenging learning tasks with action spaces of up to 17 dimensions and observation spaces of up to 376 dimensions. With these experimental results, we aim to answer three questions:
\begin{itemize}
    \item Does DACERv2 demonstrate stronger multimodal capabilities?
    \item How does the inference and training efficiency of DACERv2 compare with existing diffusion-based RL methods?
    \item How does DACERv2 compare to previous popular online RL algorithms and existing diffusion-based online RL algorithms?
\end{itemize}

\paragraph{Baselines.} The baselines encompass two categories of model-free reinforcement learning algorithms. The first category consists of diffusion-based RL methods, including a range of recent diffusion-policy online algorithms such as DACER \citep{wang2024diffusion}, QVPO \citep{ding2024diffusion}, DIME \citep{celik2025dime}, DIPO \citep{yang2023diffusion}, and QSM \citep{psenka2023learning}. The second category includes classic model-free online RL baselines, namely SAC \citep{haarnoja2018soft}, PPO \citep{schulman2017proximal}, and DSAC \citep{duan2023dsac}. The experimental hyperparameters are provided in Appendix \ref{app:hyper}. It is worth noting that the Critic network in DIME employs a two-layer MLP with a hidden dimension of 2048, consistent with their original paper, whereas the corresponding dimension for other algorithms is 256.

\paragraph{Evaluation Setups.} We implemented our algorithm in PyTorch and evaluated it on eight MuJoCo tasks using the same metrics as DACER. Experiments were conducted on a system equipped with an AMD Ryzen Threadripper 3960X 24-core processor and an NVIDIA GeForce RTX 4090 GPU. In this paper, the total training step size for all experiments was set at 1.5 million, with the results of all experiments averaged over 5 random seeds. For classic model-free baselines, we cited DACER-reported results, while all diffusion-based methods were re-evaluated. Furthermore, the training curves presented in Fig. \ref{fig:benchmark} demonstrate the stability of the training process.

\subsection{Multimodal experiments}
\label{sec: Multimodal experiments}
\begin{figure*}[ht!]
    \centering
    
    \begin{subfigure}[b]{0.23\textwidth}
        \includegraphics[width=\textwidth]{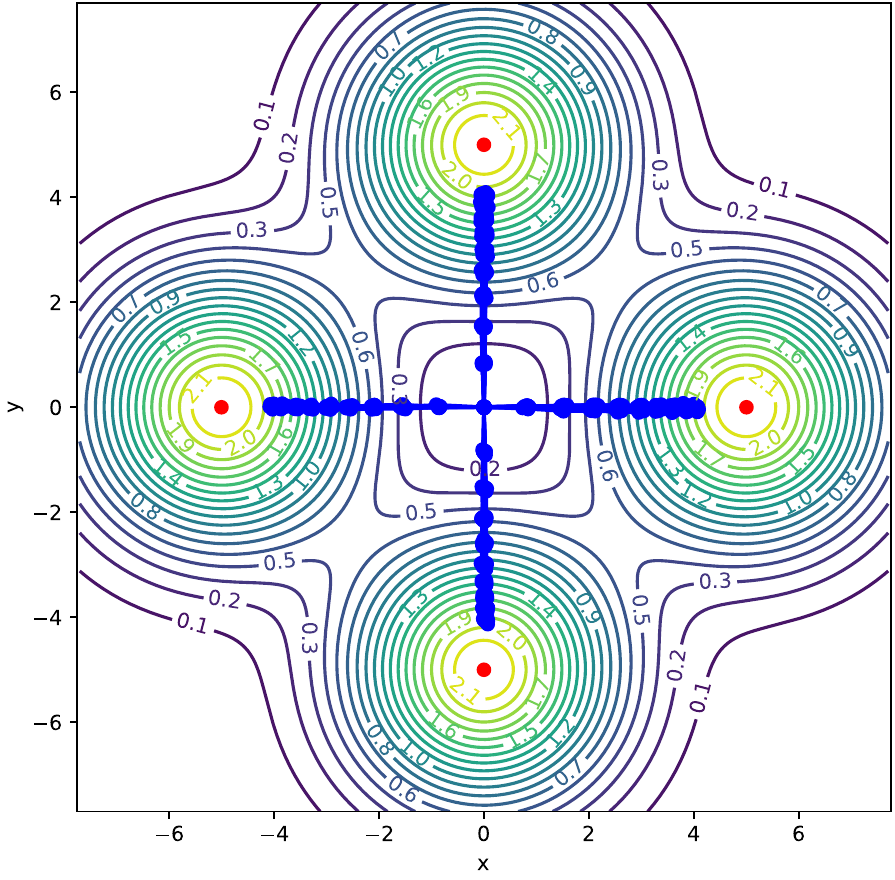}
        \caption{\textbf{DACERv2} (4 goals)}
    \end{subfigure}
    \hspace{5pt}
    \begin{subfigure}[b]{0.23\textwidth}
        \includegraphics[width=\textwidth]{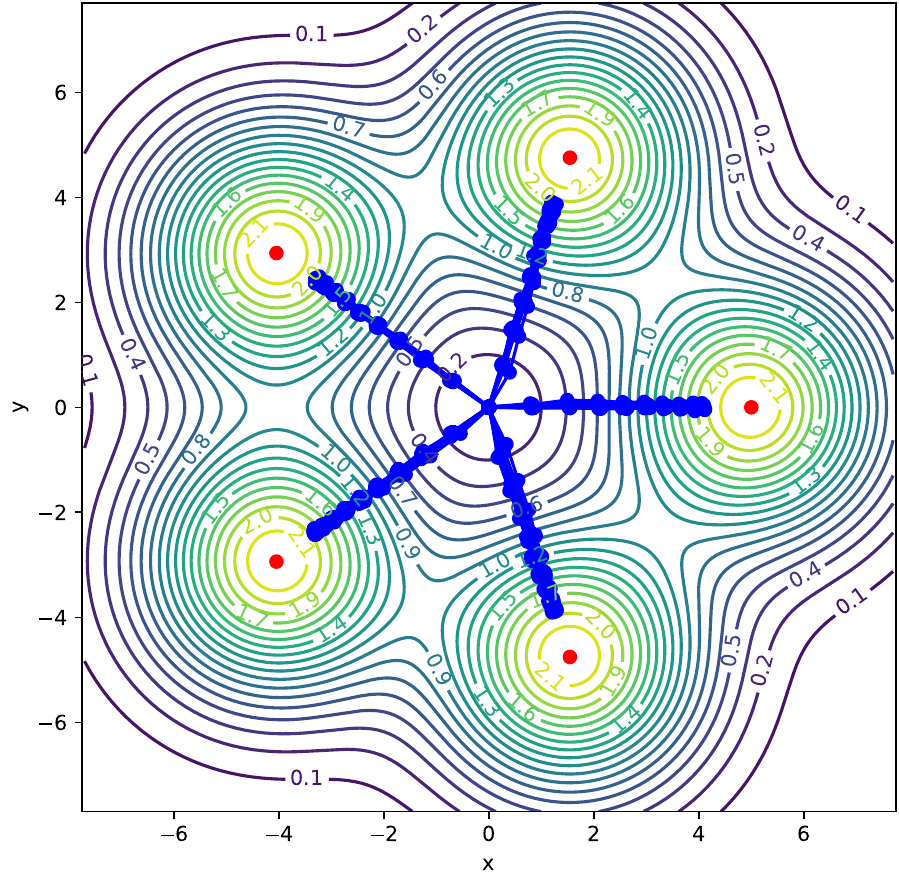}
        \caption{\textbf{DACERv2} (5 goals)}
    \end{subfigure}
    \hspace{5pt}
    \begin{subfigure}[b]{0.23\textwidth}
        \includegraphics[width=\textwidth]{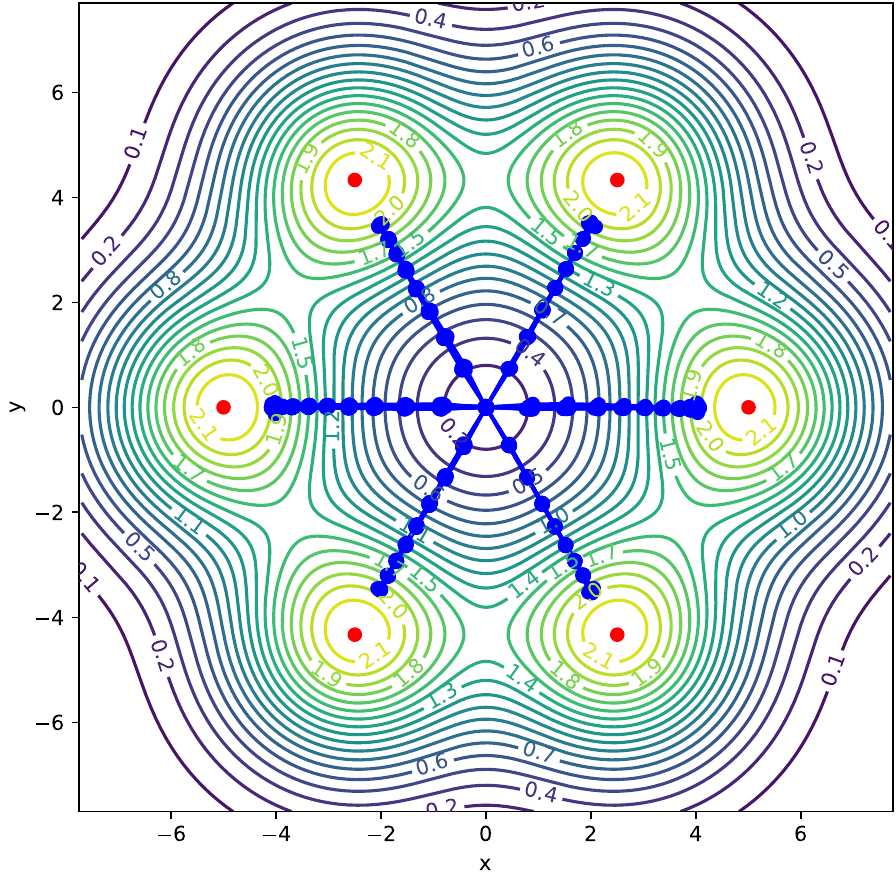}
        \caption{\textbf{DACERv2} (6 goals)}
    \end{subfigure}
    
    \begin{subfigure}[b]{0.233\textwidth}
        \includegraphics[width=\textwidth]{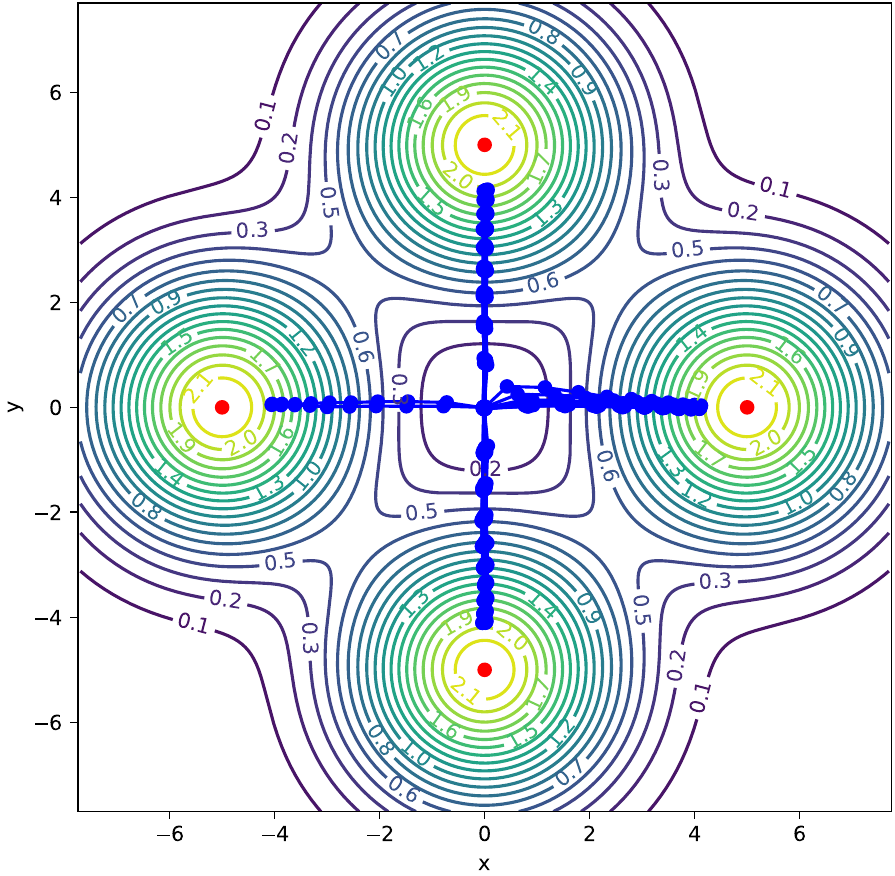}
        \caption{DACER (4 goals)}
    \end{subfigure}
    \hspace{5pt}
    \begin{subfigure}[b]{0.233\textwidth}
        \includegraphics[width=\textwidth]{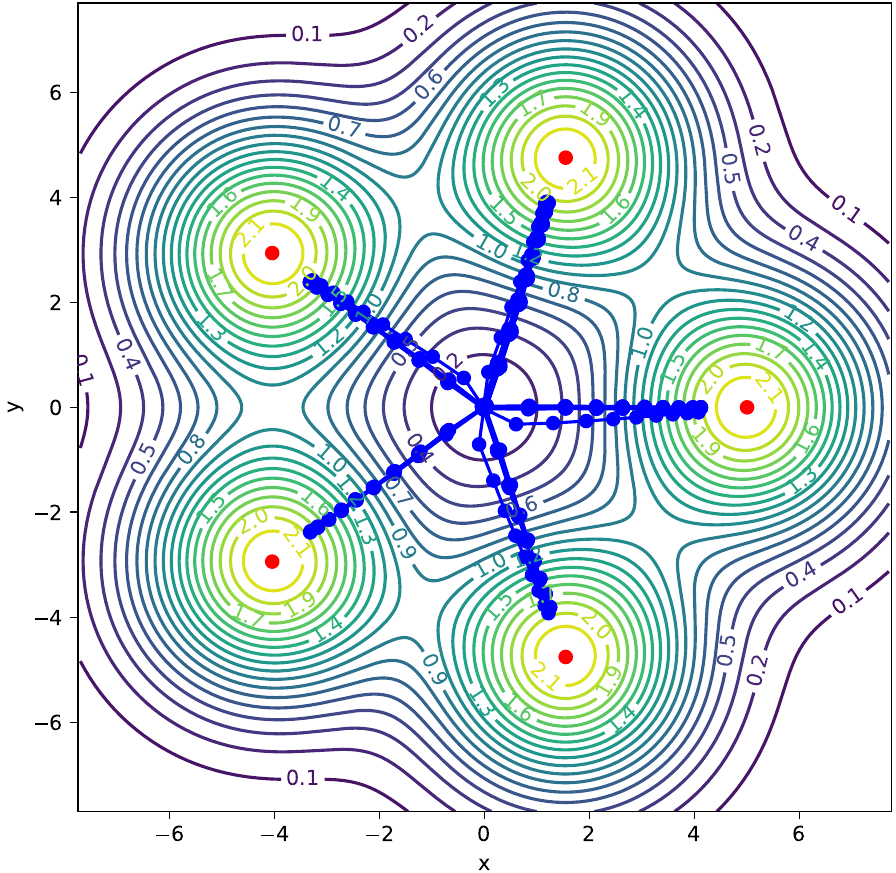}
        \caption{DACER (5 goals)}
    \end{subfigure}
    \hspace{5pt}
    \begin{subfigure}[b]{0.23\textwidth}
        \includegraphics[width=\textwidth]{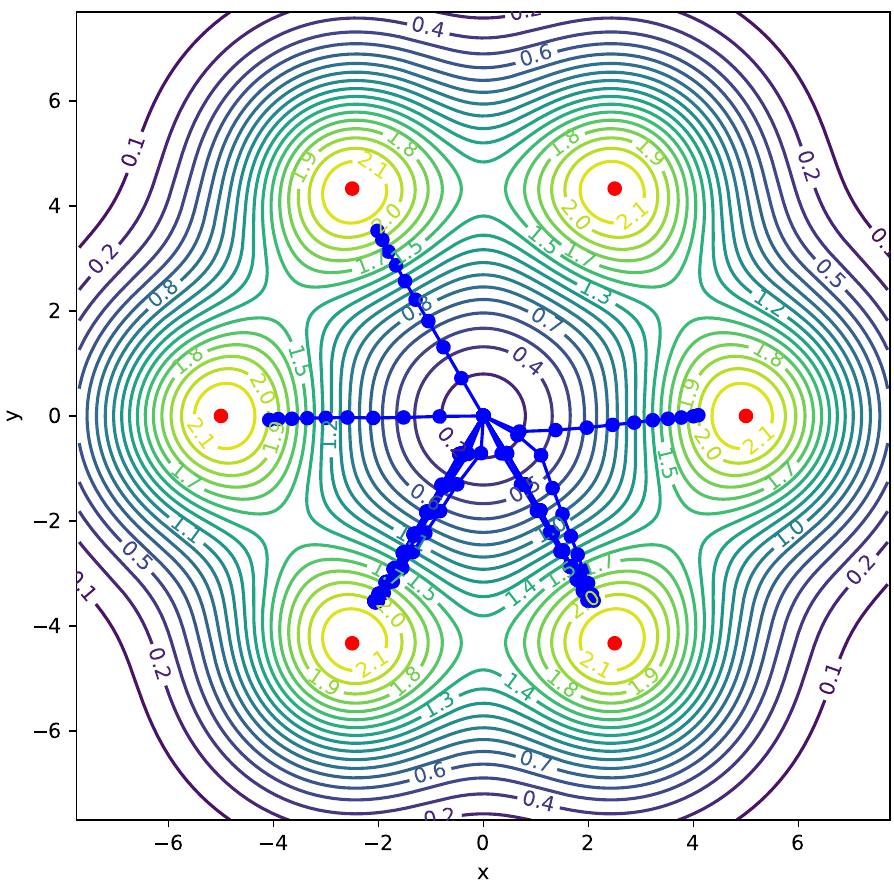}
        \caption{DACER (6 goals)}
    \end{subfigure}

    \caption{\textbf{Multi-goal Task.} Trajectories generated by policies learned using our method (top row) and original DACER (bottom row) are shown, with the $x$-axis and $y$-axis representing 2D positions (states). The agent is initialized at the origin, and the goals are marked as red dots. The level curves indicate the reward, and reaching within 1 of the endpoint signifies task completion. Results are shown for 4, 5, and 6 goal configurations from left to right.}
    \label{fig:multigoal}
\end{figure*}

We evaluate the trained policy in the ``Multi-goal'' environment by initializing the agent at the origin and sampling 100 trajectories. We conducte three sets of experiments with configurations ranging from 4 to 6 symmetrically arranged goal points. As illustrated in Fig. \ref{fig:multigoal}, the original DACER algorithm fails to maintain uniform coverage as the number of target points increases; when six targets are specified, the algorithm reaches only five target goals. In contrast, our method consistently reaches all target locations with approximately uniform coverage. These experimental results underscore that our method achieves superior exploratory capability, enabling it to more effectively capture diverse, mode-separated policies in multimodal environments.

\subsection{Efficiency Analysis}
\label{sec: Efficiency Analysis}
We first define the training time as the per-step computational cost on MuJoCo tasks, excluding the time spent on environment interaction. The inference time is measured as the latency required for the policy network to output an action given a single state as input. As illustrated in Table \ref{tab:efficiency}, the inference times of DACER, QVPO, DIME, DIPO, and QSM are \textbf{2.54×, 5.71×, 2.22×, 2.54×}, and \textbf{2.54×} longer than our method, respectively. For training time, their costs are \textbf{1.71×, 1.97×, 1.89×, 2.00×}, and \textbf{0.86×} relative to our method. Since our method achieves markedly superior performance compared to QSM, its slight disadvantage in training time is negligible in practice.
\begin{table}[ht!]
    \centering
\caption{Efficiency comparison of inference and training time. 
All values are normalized relative to \textbf{DACERv2} (set as $1.00\times$). 
Absolute times are also reported. 
Lower is better.}
    \label{tab:efficiency}
    \small
    \begin{tabular}{lcccc}
        \toprule
        \multirow{2}{*}{\textbf{Algorithms}} & 
        \multicolumn{2}{c}{\textbf{Inference Time}} & 
        \multicolumn{2}{c}{\textbf{Training Time}} \\
        \cmidrule(lr){2-3} \cmidrule(lr){4-5}
         & \textbf{Normalized} & \textbf{Absolute (ms)} & \textbf{Normalized} & \textbf{Absolute (ms)} \\
        \midrule
        \textbf{DACERv2 (Ours)} & $1.00\times$ & $0.63$ & $1.00\times$ & $7.0$ \\
        DACER   & $2.54\times$ & $1.60$ & $1.71\times$ & $12.0$ \\
        QVPO    & $5.71\times$ & $3.60$ & $1.97\times$ & $13.8$ \\
        DIME    & $2.22\times$ & $1.40$ & $1.89\times$ & $13.2$ \\
        DIPO    & $2.54\times$ & $1.60$ & $2.00\times$ & $14.0$ \\
        QSM     & $2.54\times$ & $1.60$ & $0.86\times$ & $6.0$ \\
        \bottomrule
    \end{tabular}
\end{table}

These results can be attributed to the use of a Q-gradient field objective as an auxiliary intermediate supervisory signal, which enhances the efficiency of single-step diffusion and enables our algorithm to achieve competitive performance with only five diffusion steps. Moreover, in real-time industrial control tasks, the inference time should be less than 1 milliseconds to meet control requirements. Among the existing diffusion-based algorithms, only DACERv2 (0.63 milliseconds) meets this constraint required by applications with high real-time demands.

\subsection{Experimental results}
\label{sec: comparative evaluation}
\begin{figure*}[t!]
    \centering
    \begin{subfigure}[b]{0.28\textwidth}
        \includegraphics[width=\textwidth]{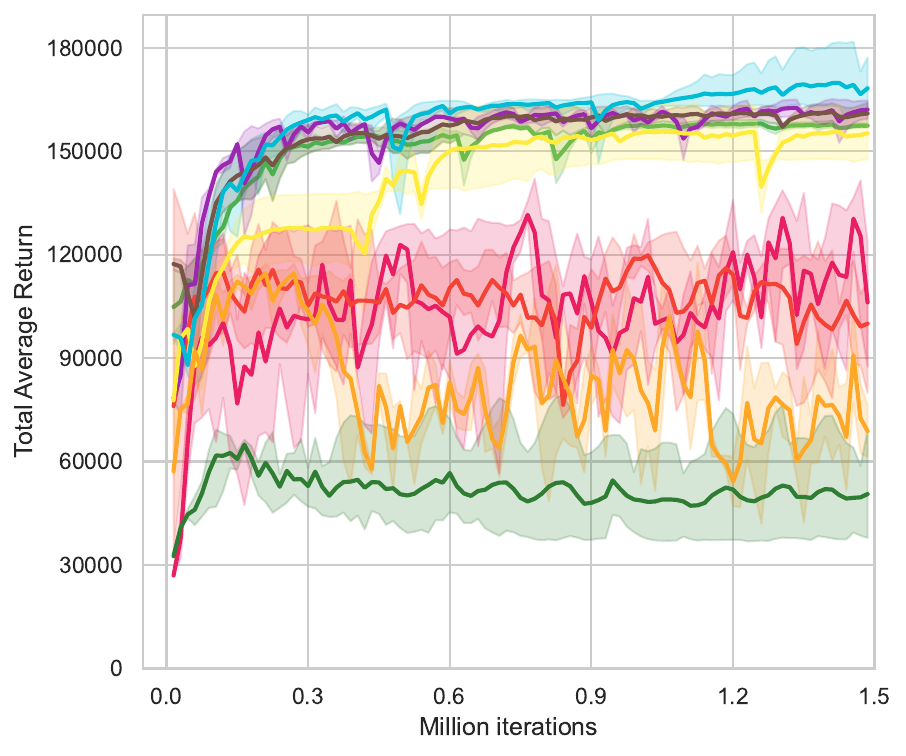}
        \caption{HumanoidStandup-v4}
        \label{fig:tar-humanoidstandup}
    \end{subfigure}
    \begin{subfigure}[b]{0.28\textwidth}
        \includegraphics[width=\textwidth]{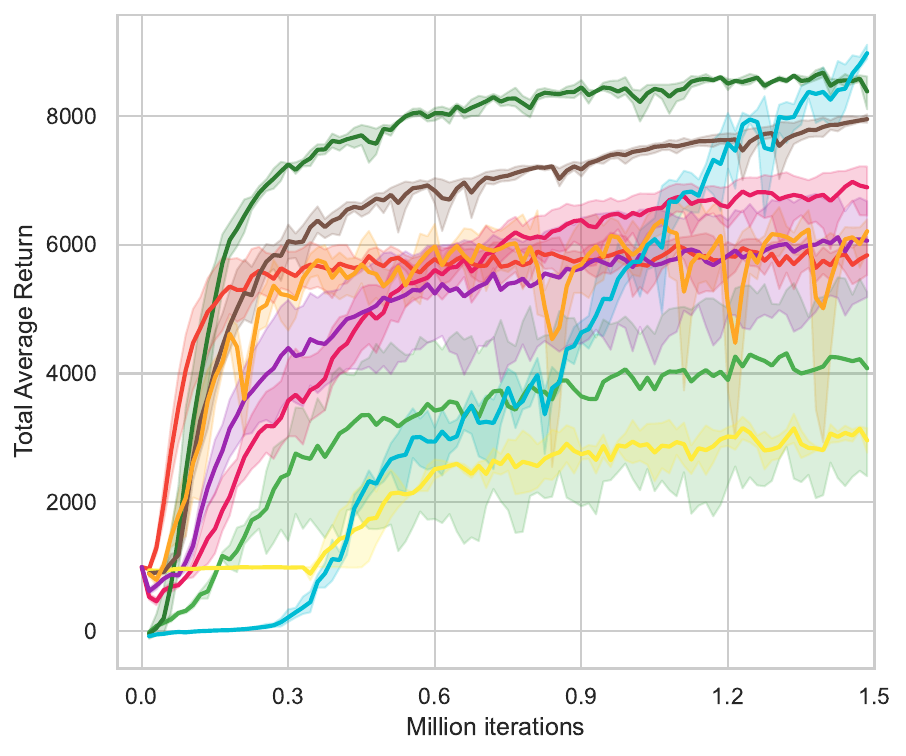}
        \caption{Ant-v3}
        \label{fig:tar-ant}
    \end{subfigure}
    \begin{subfigure}[b]{0.28\textwidth}
        \includegraphics[width=\textwidth]{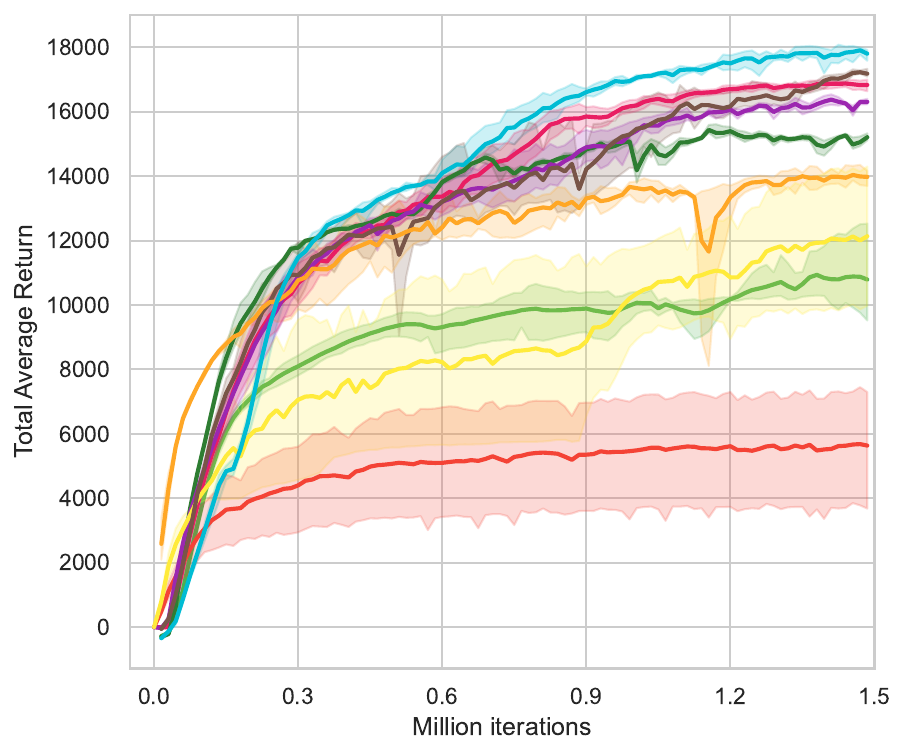}
        \caption{HalfCheetah-v3}
        \label{fig:tar-half}
    \end{subfigure}
    
    \begin{subfigure}[b]{0.28\textwidth}
        \includegraphics[width=\textwidth]{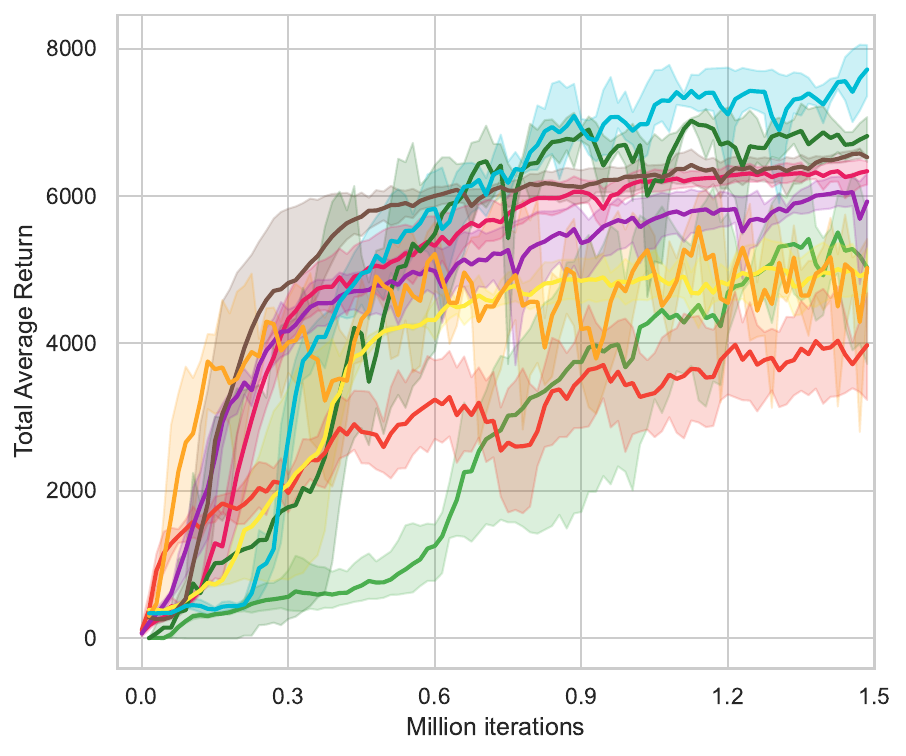}
        \caption{Walker2d-v3}
        \label{fig:tar-walker}
    \end{subfigure}
    \begin{subfigure}[b]{0.28\textwidth}
        \includegraphics[width=\textwidth]{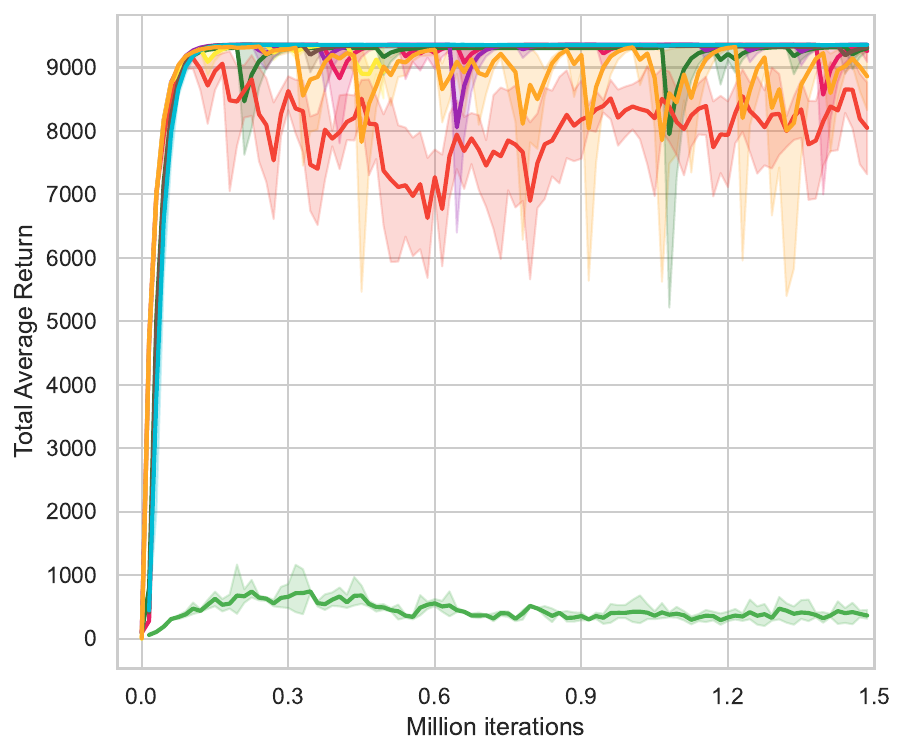}
        \caption{Inverted2Pendulum-v3}
        \label{fig:tar-inverted}
    \end{subfigure}
    \begin{subfigure}[b]{0.28\textwidth}
        \includegraphics[width=\textwidth]{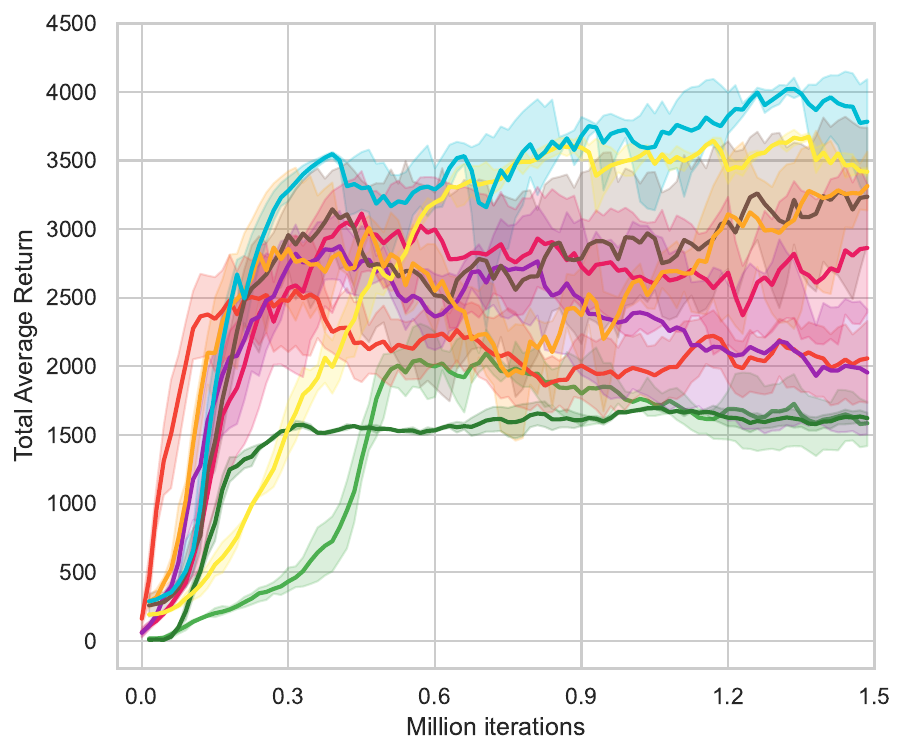}
        \caption{Hopper-v3}
        \label{fig:tar-hopper}
    \end{subfigure}
    
    \begin{subfigure}[b]{0.28\textwidth}
        \includegraphics[width=\textwidth]{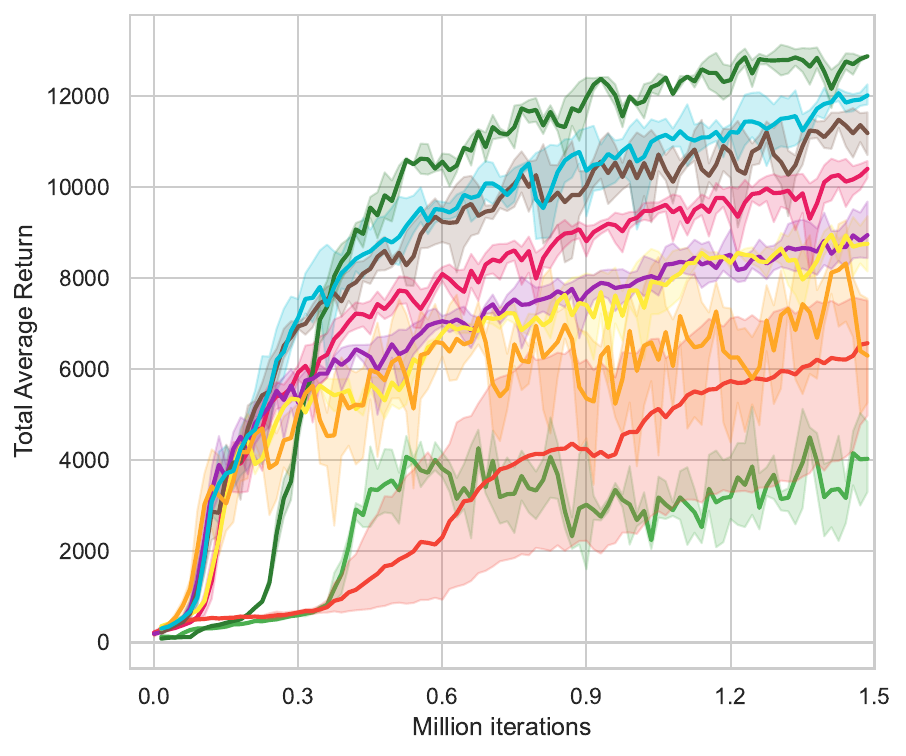}
        \caption{Humanoid-v3}
        \label{fig:tar-humanoid}
    \end{subfigure}
    \begin{subfigure}[b]{0.28\textwidth}
        \includegraphics[width=\textwidth]{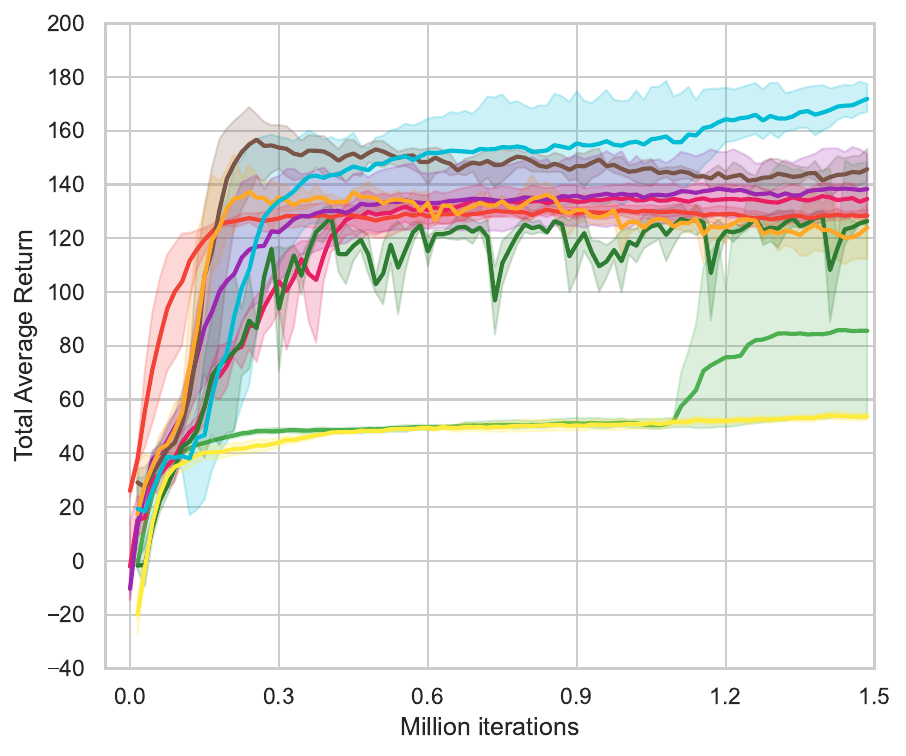}
        \caption{Swimmer-v3}
        \label{fig:tar-Swimmer}
    \end{subfigure}
    \begin{subfigure}[b]{0.28\textwidth}
        \includegraphics[width=\textwidth]{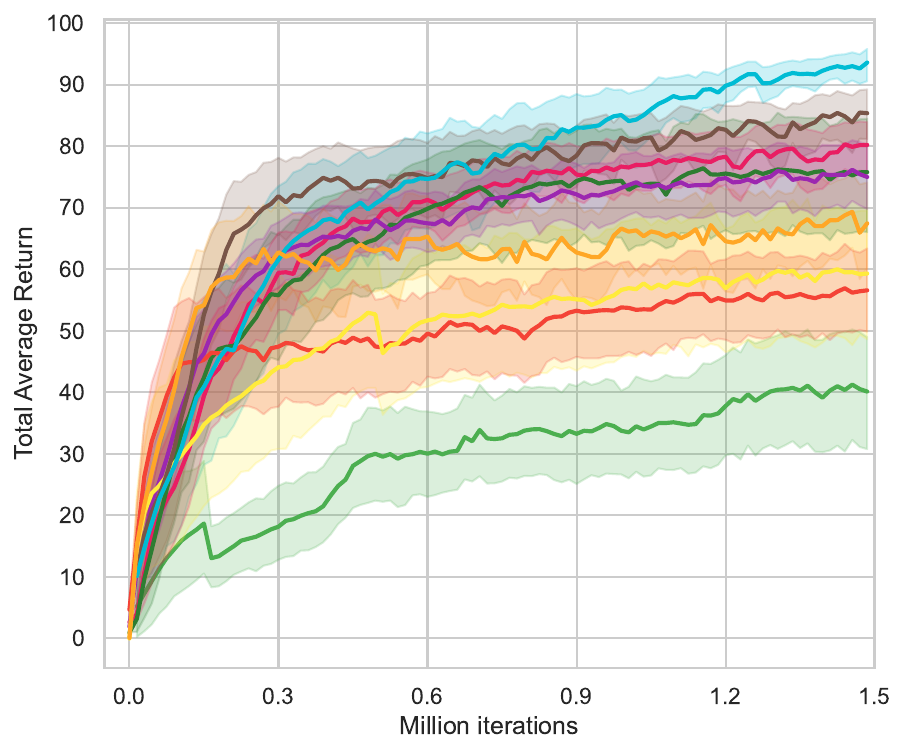}
        \caption{Normalization}
        \label{fig:tar-pusher}
    \end{subfigure}

    \includegraphics[width=0.7\linewidth]{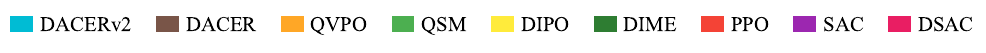}

    \caption{\textbf{Training curves on benchmarks.} 
    The solid lines represent the mean, while the shaded regions indicate the 95\% confidence interval over five runs. 
    For PPO, iterations are defined by the number of network updates.}
    \label{fig:benchmark}
\end{figure*}

All the training curves are shown in Fig. \ref{fig:benchmark} and the detailed results are listed in Table \ref{tab:benchmark_full}. Our method, DACERv2, achieves superior Total Average Return (TAR) in most complex OpenAI Gym control tasks. Despite the challenges posed by high-dimensional state and action spaces and complex dynamics, our method exhibits remarkable stability and efficiency, highlighting its robustness and adaptability.

Specifically, across challenging environments including Humanoid, Ant, HalfCheetah, HumanoidStandup, and Walker2d, our method achieves improvements of \textbf{33.1\%, 42.7\%, 9.8\%, 5.9\%}, and \textbf{29.2\%} over SAC, respectively. When compared against the best-performing diffusion-based online RL baseline in each environment, it achieves higher scores in Ant, HalfCheetah, HumanoidStandup, and Walker2d, with respective gains of \textbf{4.3\%, 4.0\%, 5.6\%}, and \textbf{10.3\%}, while underperforming DIME on Humanoid. Additionally, we normalize the returns in each task by dividing them by the highest reward across all algorithms, then average across tasks and rescale to the range of 0–100 for visualization. Under this metric, our method achieves an average score \textbf{9.7\%} higher than the second-best algorithm on the OpenAI Gym benchmark.

\begin{table*}[ht]
    \centering
    \caption{\textbf{Total Average Return (TAR).} Performance on eight tasks of OpenAI Gym MuJoCo benchmark. Mean ± Std. over 5 seeds. \textbf{Bold} = best; higher is better. The average score has been normalized to the range of 0-100.}
    \label{tab:benchmark_full}
    \renewcommand{\arraystretch}{1.4}
    \resizebox{\textwidth}{!}{%
    \begin{tabular}{lccccccccc}
    \toprule
    \textbf{Algorithm} 
      & \textbf{HumanoidStandup} 
      & \textbf{Ant} 
      & \textbf{Humanoid} 
      & \textbf{Walker2d} 
      & \textbf{Inverted2Pendulum} 
      & \textbf{Hopper} 
      & \textbf{HalfCheetah} 
      & \textbf{Swimmer} 
      & \textbf{Average score} \\
    \midrule
    PPO 
      & $82807 \pm 8633$ 
      & $6157 \pm 185$ 
      & $6869 \pm 1563$ 
      & $4832 \pm 638$ 
      & $9357 \pm 2$ 
      & $2647 \pm 481$ 
      & $5789 \pm 2201$ 
      & $130 \pm 2$ 
      & $56.69 \pm 19.80$ \\
    SAC 
      & $161413 \pm 1643$ 
      & $6427 \pm 804$ 
      & $9335 \pm 696$ 
      & $6201 \pm 263$ 
      & $\mathbf{9360 \pm 0}$ 
      & $2483 \pm 943$ 
      & $16573 \pm 224$ 
      & $140 \pm 14$ 
      & $75.41 \pm 17.64$ \\
    DSAC 
      & $149576 \pm 1795$ 
      & $7086 \pm 261$ 
      & $10829 \pm 243$ 
      & $6424 \pm 147$ 
      & $\mathbf{9360 \pm 0}$ 
      & $3660 \pm 533$ 
      & $17025 \pm 157$ 
      & $138 \pm 6$ 
      & $80.18 \pm 12.07$ \\
    QSM 
      & $150692 \pm 1497$ 
      & $4783 \pm 1235$ 
      & $6072 \pm 691$ 
      & $5685 \pm 437$ 
      & $591 \pm 98$ 
      & $2006 \pm 251$ 
      & $11401 \pm 882$ 
      & $46 \pm 1$ 
      & $44.54 \pm 25.05$ \\
    DIPO 
      & $156870 \pm 8270$ 
      & $3449 \pm 149$ 
      & $9353 \pm 356$ 
      & $5066 \pm 365$ 
      & $9355 \pm 2$ 
      & $3813 \pm 241$ 
      & $12267 \pm 2180$ 
      & $55 \pm 2$ 
      & $63.93 \pm 23.00$ \\
    DIME 
      & $78303 \pm 3165$ 
      & $8789 \pm 105$ 
      & $\mathbf{13065 \pm 221}$ 
      & $7261 \pm 299$ 
      & $9356 \pm 2$ 
      & $2016 \pm 179$ 
      & $15816 \pm 292$ 
      & $134 \pm 3$ 
      & $75.87 \pm 22.34$ \\
    DACER 
      & $161928 \pm 3804$ 
      & $8040 \pm 128$ 
      & $11791 \pm 238$ 
      & $6674 \pm 169$ 
      & $9354 \pm 2$ 
      & $4062 \pm 181$ 
      & $17488 \pm 216$ 
      & $150 \pm 4$ 
      & $84.98 \pm 11.38$ \\
    QVPO 
      & $129865 \pm 8932$ 
      & $6484 \pm 145$ 
      & $9656 \pm 252$ 
      & $6057 \pm 352$ 
      & $9354 \pm 5$ 
      & $4035 \pm 172$ 
      & $14355 \pm 175$ 
      & $130 \pm 10$ 
      & $67.80 \pm 16.74$ \\
    \textbf{DACERv2 (ours)} 
      & $\mathbf{170956 \pm 8792}$ 
      & $\mathbf{9169 \pm 129}$ 
      & $12426 \pm 292$ 
      & $\mathbf{8011 \pm 188}$ 
      & $9359 \pm 1$ 
      & $\mathbf{4202 \pm 191}$ 
      & $\mathbf{18192 \pm 266}$ 
      & $\mathbf{172 \pm 6}$ 
      & $\mathbf{93.19 \pm 6.35}$ \\
    \bottomrule
    \end{tabular}%
    }
\end{table*}

\subsection{Ablation Study}
\label{sec:ablation}
In this section, we conduct ablation study to investigate the impact of the following four aspects on the performance of the diffusion policy: 1) whether to use the Q-gradient field training objective function; 2) whether to use time-weighted mechanism; 3) different diffusion step size $T$; 4) the sensitivity to the hyperparameter $\eta$. The experiments are conducted in the Walker2d-v3 task. Ablation study on the effect of Q-value normalization is provided in Appendix \ref{app: extra ablation study}.

\begin{figure}[ht!]
\centering
  \begin{subfigure}{0.3\linewidth} 
    \centering
    \includegraphics[width=\textwidth]{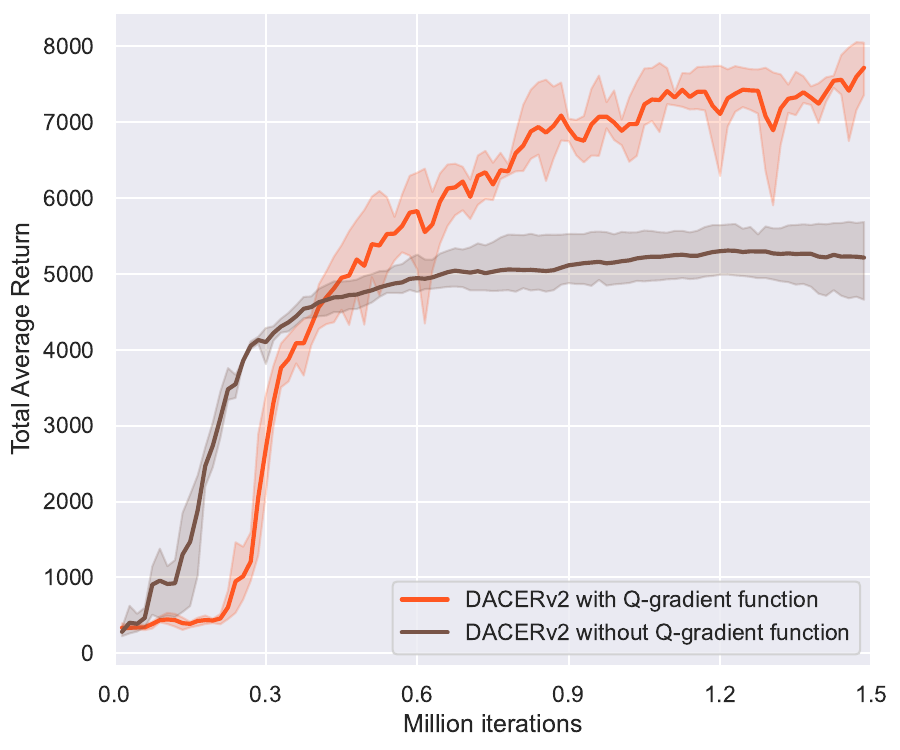} 
    \label{fig:DACERv1-v2}
    \caption{\centering{Ablation for the Q-gradient field training objective function.}}
  \end{subfigure}
  \begin{subfigure}{0.3\linewidth}
    \centering
    \includegraphics[width=\textwidth]{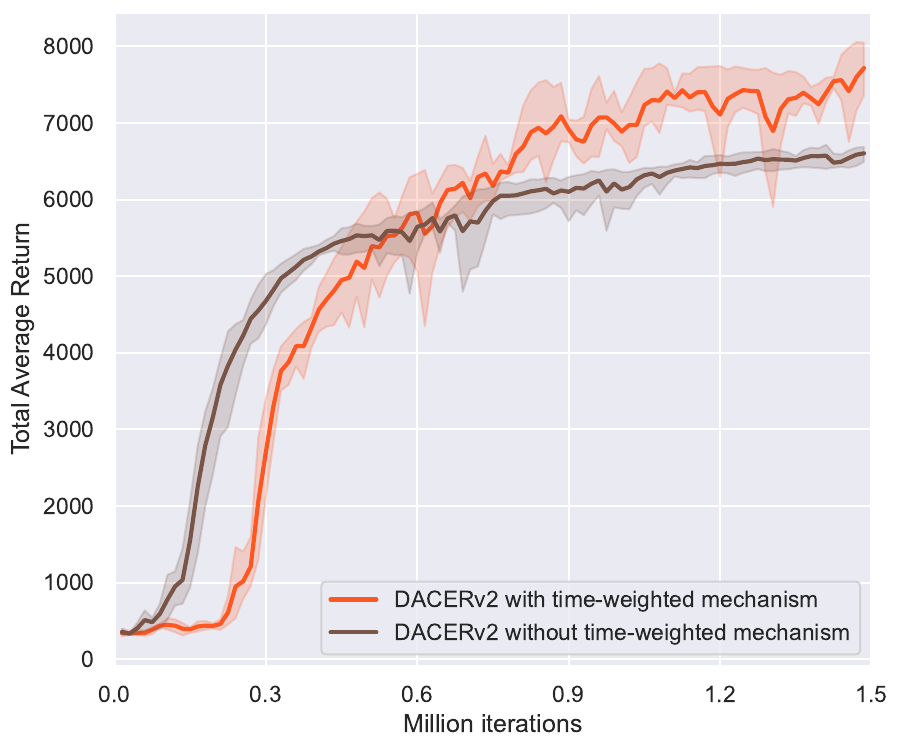} 
    \label{fig:rm_TAR}
    \caption{\centering{Ablation for the time-weighted mechanism.}}
  \end{subfigure}
  \begin{subfigure}{0.3\linewidth}
    \centering
    \includegraphics[width=\textwidth]{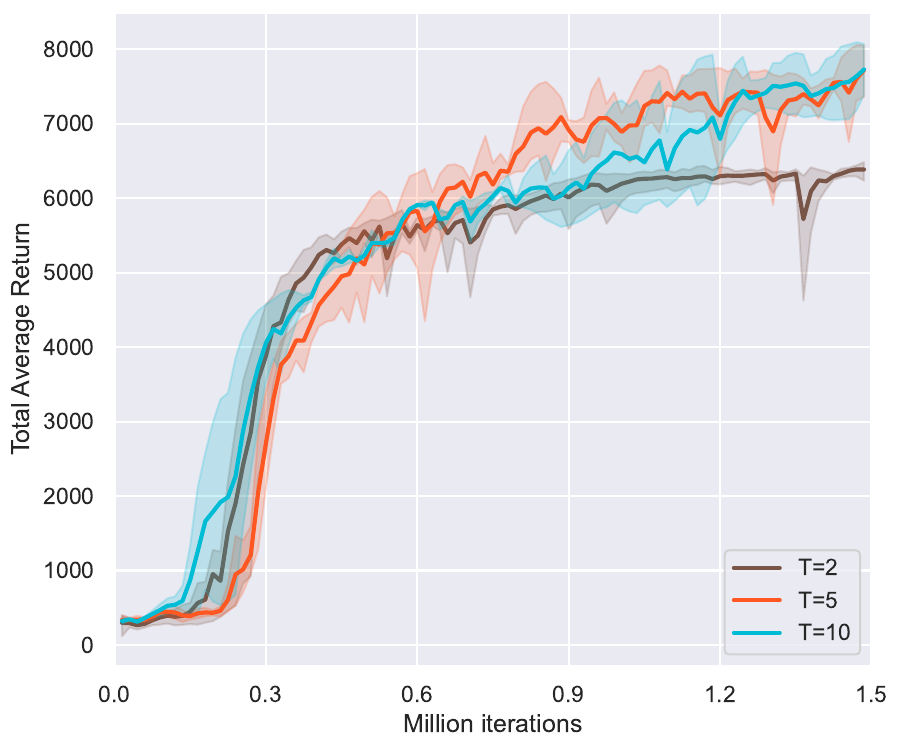} 
    \label{fig:rm_TAR}
    \caption{\centering{Ablation for the different diffusion steps.}}
  \end{subfigure}
  
\caption{\textbf{Ablation experiment curves.} (a) The performance of DACERv2 with Q-gradient function on Walker2d-v3 is far better than without Q-gradient function. (b) Time-weighted mechanism can further improve the performance of our algorithm. (c) A diffusion step size of 5 provides a balance between efficiency and performance.} 
\label{fig: DACERv2 Ablation}
\end{figure}

\paragraph{Q-gradient field training objective function.} In this ablation study, we fixed the diffusion step size at 5 to examine the effect of incorporating the Q-gradient field loss function. As shown in Fig.~\ref{fig: DACERv2 Ablation}(a), removing this objective caused a substantial drop in performance. This finding highlights the critical role of the Q-gradient field loss in guiding the diffusion denoising process and demonstrates its importance as a key component for enhancing overall performance.

\paragraph{Time-weighted mechanism.} We conducted an experiment to demonstrate that using time-weighted mechanism can further improve performance. As shown in Fig. \ref{fig: DACERv2 Ablation}(b), directly using $\nabla_{a}Q(s, a)$ as the target value in the Q-gradient field training loss, instead of the $w(t)\nabla_{a}Q(s, a)$, results in performance degradation. This is because different timesteps require matching different magnitudes of noise prediction, which enhances both training stability and final performance.

\paragraph{Diffusion steps.} We further investigated the performance of the diffusion policy under varying numbers of diffusion timesteps $T$. We plotted training curves for $T = 2, 5,$ and $10$, as shown in Fig.~\ref{fig: DACERv2 Ablation}(c). The experimental results suggest that increasing the number of diffusion steps does not necessarily improve performance, while using fewer steps tends to degrade performance.

\paragraph{The sensitivity to the hyperparameter $\eta$.} To assess the sensitivity of $\eta$, we evaluated five settings ($0.1, 0.01, 0.001, 0.012, 0.008$) on Humanoid-v3. As reported in Table~\ref{tab:eta_performance_comparison}, performance degraded markedly at $\eta=0.1$ and $0.001$, but remained stable at $\eta=0.012$ and $0.008$, indicating tenfold sensitivity. These results suggest that the algorithm is robust to moderate variations in $\eta$ and thus does not require extensive hyperparameter tuning.

\begin{table}[ht]
    \centering
    \caption{Performance comparison of DACERv2 with different $\eta$ values on Humanoid-v3.}
    \label{tab:eta_performance_comparison}
    \small
    \begin{tabular}{@{}lccccc@{}}
        \toprule
        \textbf{Algorithm} & $\eta=0.01$ & $\eta=0.1$ & $\eta=0.001$ & $\eta=0.012$ & $\eta=0.008$ \\
        \midrule
        \textbf{DACERv2} & $\mathbf{12426 \pm 292}$ & $11463 \pm 304$ & $11161 \pm 287$ & $12101 \pm 325$ & $12208 \pm 249$ \\
        \bottomrule
    \end{tabular}
\end{table}


\section{Conclusion}
\label{sec:conclusion}
In this paper, we address the critical challenge of balancing performance and time-efficiency in diffusion-based online RL. By introducing a Q-gradient field objective and a time-dependent weighting scheme, our method enables each denoising step to be guided by the Q-function with adaptive emphasis over time. This design allows the policy to achieve strong performance with only five diffusion steps, significantly improving both training and inference speed. 
\bibliography{ref}
\bibliographystyle{iclr2026_conference}

\appendix
\newpage

\section{Theoretical analysis}
\label{app: Additional explanation}
\begin{theorem}
Let $\mathcal{S}$ denote the state space and $\mathcal{A}$ denote the continuous action space. Suppose $p(s)$ is a distribution over states, $\mathcal{H}_0^{global}$ denotes a specific entropy value. We define the policy space $\Pi_{\mathcal{H}_0^{global}}$ as the set of policy families $\{\pi^*(\cdot|s)\}_{s \in \mathcal{S}}$, where each $\pi(\cdot|s)$ represents a valid probability distribution over actions. This policy family is required to satisfy a global expected entropy constraint:
\begin{equation}
\mathbb{E}_{s \sim p(s)}[H(\pi^*(\cdot|s))] = \mathcal{H}_0^{global},
\end{equation}
where $\mathcal{H}_0^{global}$ is a given constant.

Within the policy space $\Pi_{\mathcal{H}_0^{global}}$, the family of policies $\{\pi^*(\cdot|s)\}_{s \in \mathcal{S}}$ that maximizes the global expected action value $\mathbb{E}_{s \sim p(s)}[\mathbb{E}_{a \sim \pi(a|s)}[Q(s,a)]$ has the property that, for each state $s$, the optimal policy $\pi^*(a|s)$ takes the form of a soft policy:
\begin{equation}
\pi^*(a|s) = \frac{\exp(Q(s,a)/\alpha)}{\int_{a' \in \mathcal{A}} \exp(Q(s,a')/\alpha) da'},
\end{equation}
where $\alpha>0$ is a global temperature parameter, whose value is implicitly determined by a global expected entropy constraint: $\mathbb{E}_{s \sim p(s)}[H(\pi^*(\cdot|s))] = \mathcal{H}_0^{global}$.
\label{the: soft policy}
\end{theorem}

\noindent \textit{\textbf{Proof.}} We seek a family of policies $\{\pi(\cdot\mid s)\}_{s\in\mathcal S}$ belonging to the constrained space:
\begin{equation}
   \Pi_{\mathcal H^{\text{global}}_0}\;=\;\Bigl\{\{\pi(\cdot\mid s)\}_{s\in\mathcal S}\ \Bigm|\ 
\mathbb{E}_{s\sim p(s)}\bigl[H(\pi(\cdot\mid s))\bigr]=\mathcal H^{\text{global}}_0,\;
\int_{\mathcal A}\!\pi(a\mid s)\,da=1,\ \forall s\Bigr\}, 
\end{equation}
which maximises the expected action-value
\begin{equation}
   J\bigl(\{\pi(\cdot\mid s)\}\bigr)
\;=\;
\mathbb{E}_{s\sim p(s)}\!\Bigl[\mathbb{E}_{a\sim\pi(\cdot\mid s)}\!\bigl[Q(s,a)\bigr]\Bigr]
\;=\;
\int_{\mathcal S}p(s)\!\!\int_{\mathcal A}\pi(a\mid s)\,Q(s,a)\,da\,ds . 
\end{equation}

Then, we introduce a scalar multiplier $\alpha$ for the global expected-entropy constraint and a state-dependent multiplier $\eta(s)$ for the normalisation constraint at each $s$.
The Lagrangian reads

\begin{equation}
    \begin{aligned}
\mathcal L\bigl(\{\pi(\cdot\mid s)\},\alpha,\{\eta(s)\}\bigr)
&=
\int_{\mathcal S}\!\int_{\mathcal A}
\Bigl[p(s)\pi(a\mid s)Q(s,a)
-\alpha\,p(s)\pi(a\mid s)\log\pi(a\mid s)
+\eta(s)\pi(a\mid s)\Bigr]\,da\,ds
\\
&\quad
-\alpha\,\mathcal H^{\text{global}}_0
-\int_{\mathcal S}\eta(s)\,ds .
\end{aligned}
\end{equation}

Because the decision variables for distinct states couple only through $\alpha$, we can minimise the integrand for each fixed $s$ independently:
\begin{equation}
    \mathcal L_s\bigl(\pi(\cdot\mid s)\bigr)
\;=\;
\int_{\mathcal A}
\Bigl[p(s)\pi(a\mid s)Q(s,a)
-\alpha\,p(s)\pi(a\mid s)\log\pi(a\mid s)
+\eta(s)\pi(a\mid s)\Bigr]\,da .
\end{equation}

Taking the functional derivative and setting it to zero yields, for almost every $a\in\mathcal A$, we can obtain
\begin{equation}
p(s)Q(s,a)\;-\;\alpha\,p(s)\log\pi(a\mid s)\;-\;\alpha\,p(s)\;+\;\eta(s)\;=\;0 .
\end{equation}

Assuming $p(s)>0$, we divide both sides by $p(s)$ and rearrange:
\begin{equation}
    \log\pi(a\mid s)
\;=\;\frac{Q(s,a)}{\alpha}\;-\;1\;+\;\frac{\eta(s)}{\alpha p(s)}.
\end{equation}

Let $\tilde\eta(s)=\eta(s)/p(s)$.  Exponentiating gives the unnormalised form
\begin{equation}
  \pi(a\mid s)
\;=\;
\exp\!\Bigl(\frac{\tilde\eta(s)-\alpha}{\alpha}\Bigr)
\exp\!\Bigl(\tfrac{Q(s,a)}{\alpha}\Bigr)
\;=\;
C(s)\,\exp\!\Bigl(\tfrac{Q(s,a)}{\alpha}\Bigr),  
\end{equation}
where $C(s)$ is a state-wise normalising constant.

Imposing $\int_{\mathcal A}\pi(a\mid s)\,da=1$, we can determine
\begin{equation}
    C(s)
=
\Bigl[\int_{\mathcal A}\!\exp\!\bigl(Q(s,a')/\alpha\bigr)\,da'\Bigr]^{-1}.
\end{equation}

Therefore, the optimal policy family is the Boltzmann distribution
\begin{equation}
  \pi^{*}(a\mid s)
=
\frac{\exp\!\bigl(Q(s,a)/\alpha\bigr)}
{\displaystyle\int_{\mathcal A}\!\exp\!\bigl(Q(s,a')/\alpha\bigr)\,da'}
\qquad\forall s\in\mathcal S,\;a\in\mathcal A .  
\end{equation}

The scalar $\alpha>0$ is the Lagrange multiplier associated with the global entropy constraint and serves as a common temperature across all states.  Its value is obtained implicitly by substituting $\pi^{*}$ back into
\begin{equation}
  \mathbb{E}_{s\sim p(s)}\!\bigl[H(\pi^{*}(\cdot\mid s))\bigr]
=\mathcal H^{\text{global}}_0 .  
\end{equation}

Consequently, although the entropy constraint is imposed only on the state-averaged entropy, each per-state optimal policy still follows a Boltzmann form with the same temperature parameter $\alpha$.

\section{Related Work}
\label{sec:Related work}
We review existing works on using the diffusion model as a policy function in combination with RL.

\paragraph{Online RL with Diffusion Policy.} Online RL enables agents to refine their policies through real-time interaction. Yang \textit{et al.} introduced DIPO \citep{yang2023policy}, which maintains a dedicated diffusion buffer to store actions and model them using diffusion techniques. Psenka \textit{et al.} proposed QSM \citep{psenka2023learning}, which aligns policies with $\nabla_a Q$ via score matching, but is sensitive to value gradient inaccuracies across the action space. Recently, Ding \textit{et al.} \citep{ding2024diffusion} proposed QVPO, which weights diffusion-sampled actions by Q-values without computing gradients. However, it uses a fixed ratio of uniform samples to boost the entropy, lacking adaptive control and later degrading performance. Ma \textit{et al.} \citep{ma2025soft} proposed SDAC, which uses score matching over noisy energy-based diffusion. It avoids requiring optimal actions but suffers from high gradient variance due to poor sampling in high-Q regions. Celik \textit{et al.} proposed DIME \citep{celik2025dime}, which derives a lower bound on the diffusion policy entropy and integrates it into the maximum-entropy RL framework. However, directly incorporating an inaccurate entropy estimate into the policy objective can degrade performance.

\paragraph{Offline RL with Diffusion Policy.} Offline RL focuses on learning optimal policies from suboptimal datasets, with the core challenge being the out-of-distribution (OOD) problem~\citep{kumar2020conservative, fujimoto2019off}. Diffusion models are naturally suited for offline RL due to their ability to model complex data distributions. Wang \textit{et al.} proposed Diffusion-QL~\citep{wang2022diffusion}, which combines behavior cloning through a diffusion loss with Q-learning to improve policy learning. However, Diffusion-QL suffers from slow training and instability in OOD regions. To address the former, Kang \textit{et al.} proposed Efficient Diffusion Policy (EDP)~\citep{kang2023efficient}, which speeds up training by initializing from dataset actions and adopting a one-step sampling strategy. To mitigate OOD issues, Ada \textit{et al.} introduced SRDP~\citep{ada2024diffusion}, which enhances generalization by integrating state reconstruction into the diffusion policy. Furthermore, Chen \textit{et al.} proposed CPQL~\citep{chen2023boosting}, a consistency-based method that improves efficiency via one-step noise-to-action generation during both training and inference, albeit with some performance trade-offs.

\paragraph{Comparison with DACER.} Wang \textit{et al.} proposed DACER \citep{wang2024diffusion}, which uses the reverse diffusion process as a policy approximator and employs a GMM to estimate policy entropy for balancing exploration and exploitation. However, the work lacks a theoretical justification for why maximizing the expected Q-value enables learning multimodal policies. Moreover, a key trade-off remains: long diffusion processes hinder training efficiency, while fewer steps lead to performance degradation. Our work addresses this by reducing the diffusion steps while maintaining or even improving both performance and policy multimodality.

\paragraph{Comparison with QSM.} Psenka \textit{et al.} proposed QSM \citep{psenka2023learning}, an algorithm that aligns diffusion model policies with $\nabla_{\boldsymbol{a}} Q(\boldsymbol{s}, \boldsymbol{a})$ by leveraging their score-based structure. Both methods leverage Q-gradients for diffusion policy optimization. QSM employs score matching, whereas DACERv2 performs end-to-end Q-value maximization augmented with a time-weighted score-matching loss and entropy regularization, resulting in a multi-task objective. DACERv2 additionally stabilizes Q-gradients through normalization and improves efficiency, converging in just 5 diffusion steps compared to QSM’s approximately 20.

\section{Environmental Details}
\label{app:env details}
\paragraph{MuJoCo \citep{2012MuJoCo}:} This is a high-performance physics simulation platform widely adopted for robotic reinforcement learning research. The environment features efficient physics computation, accurate dynamic system modeling, and comprehensive support for articulated robots, making it an ideal benchmark for RL algorithm development. In this research, we concentrate on eight tasks: Humanoid, Ant, HalfCheetah, Walker2d, InvertedDoublePendulum (IDP), Hopper, HumanoidStandup, and Swimmer. The IDP task entails maintaining the balance of a double pendulum in an inverted state. In contrast, the objective of the other tasks is to maximize the forward velocity while avoiding falling. All these tasks are realized through the OpenAI Gym interface \citep{brockman2016openai}. 

\begin{figure}[h!]

    \begin{minipage}[b]{0.25\textwidth}
        \centering
        \includegraphics[width=\linewidth]{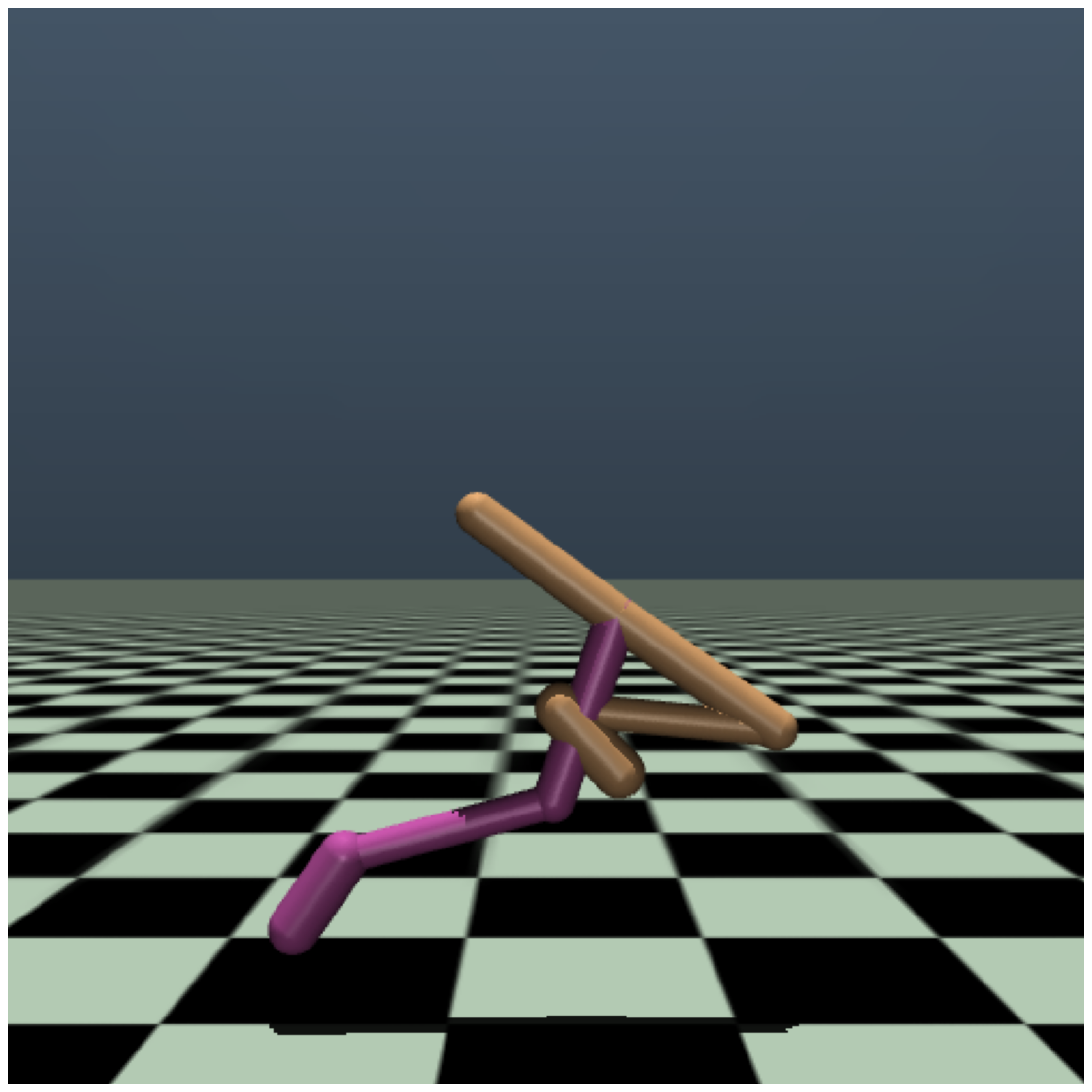}
        \caption{Walker2d-v3}
        \label{fig:walker2d}
    \end{minipage}%
    \quad
    \begin{minipage}[b]{0.6\textwidth}
    \small
        \textbf{\textit{State-action space}}: $\mathcal{S} \in \mathbb{R}^{17}$, $\mathcal{A} \in \mathbb{R}^{6}$.\\
        
        \textbf{\textit{Objective.}} Maintain forward velocity as fast as possible while avoiding falling over.\\
        
        \textbf{\textit{Initialization.}} The walker is initialized in a standing position with slight random noise added to joint positions and velocities.\\
        
        \textbf{\textit{Termination.}} The episode ends when the agent falls, the head touches the ground, or after 1000 steps.\\
    \end{minipage}
\end{figure}

\begin{figure}[h!]
\small
    \begin{minipage}[b]{0.25\textwidth}
        \centering
        \includegraphics[width=\linewidth]{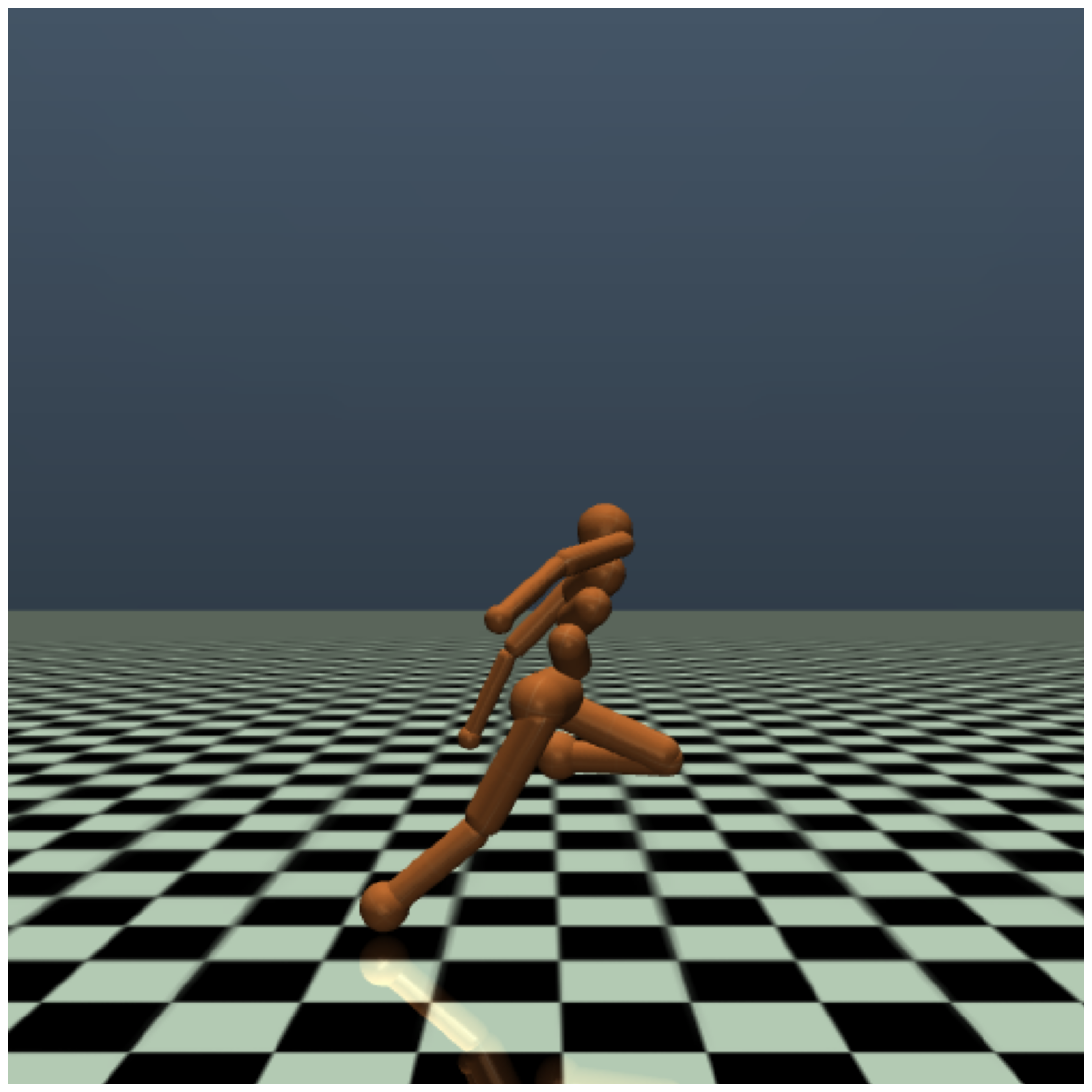}
        \caption{Humanoid-v3}
        \label{fig:humanoid}
    \end{minipage}%
    \quad
    \begin{minipage}[b]{0.6\textwidth}
\textbf{\textit{State-action space}}: $\mathcal{S} \in \mathbb{R}^{376}$, $\mathcal{A} \in \mathbb{R}^{17}$.\\

        \textbf{\textit{Objective.}} Maintain balance and walk or run forward at a high velocity while avoiding falls.\\
        
\textbf{\textit{Initialization.}} The humanoid starts in an upright position with slight random perturbations to joint angles and velocities.\\

\textbf{\textit{Termination.}} The episode ends when the head height is less than 1.0 meter, the torso tilts excessively, or after 1000 steps.\\
       
    \end{minipage}
\end{figure}

\begin{figure}[h!]
    \begin{minipage}[b]{0.4\textwidth}
        \centering
        \includegraphics[width=\linewidth]{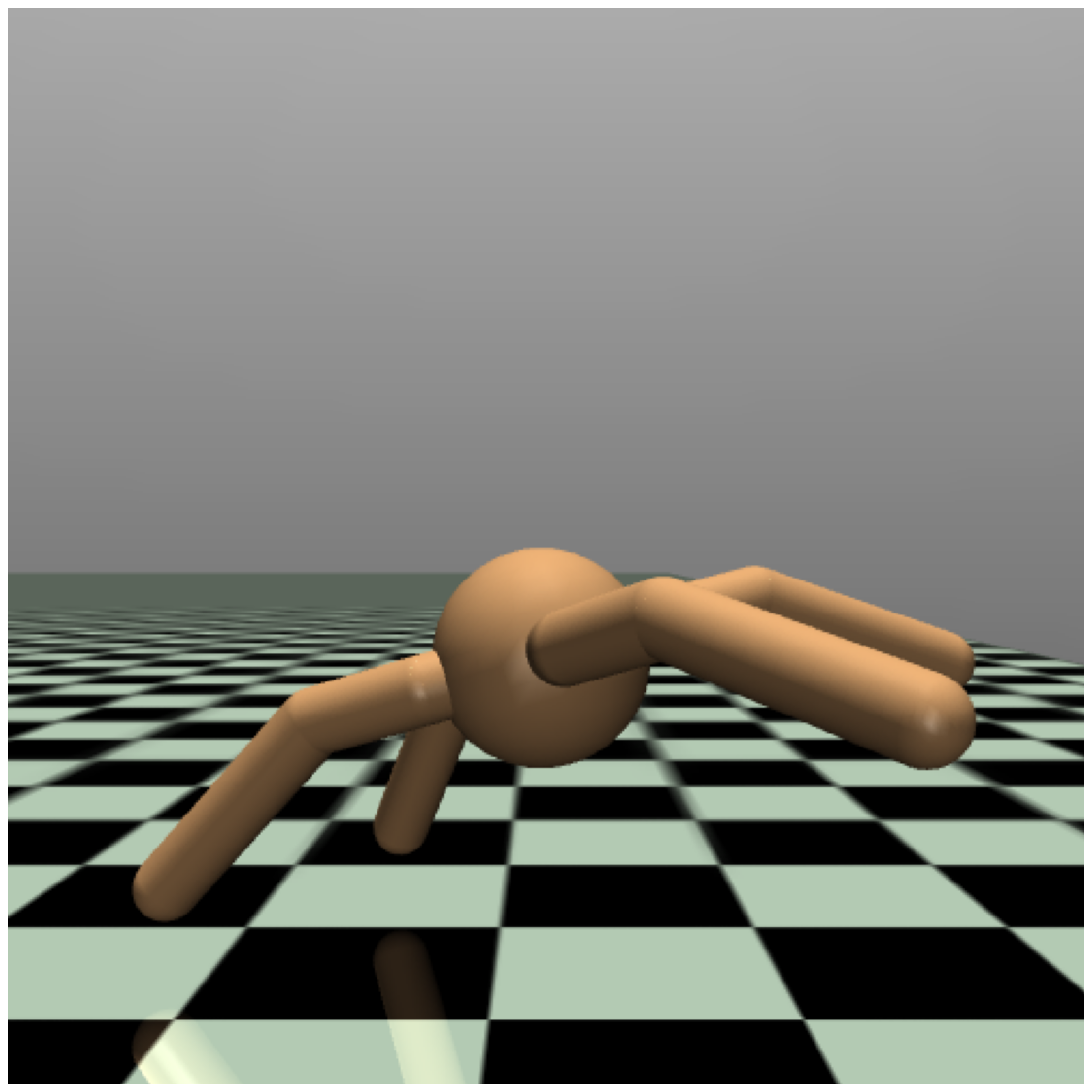}
        \caption{Ant-v3}
        \label{fig:ant}
    \end{minipage}%
    \quad
    \begin{minipage}[b]{0.6\textwidth}
    \small
\textbf{\textit{State-action space}}: $\mathcal{S} \in \mathbb{R}^{111}$, $\mathcal{A} \in \mathbb{R}^{8}$.\\

\textbf{\textit{Objective.}} Navigate forward as quickly as possible using four legs while maintaining stability.\\

\textbf{\textit{Initialization.}} The ant is initialized in a stable, upright position with random noise applied to its joints.\\

\textbf{\textit{Termination.}} The episode ends if the ant falls, flips over, or reaches the maximum step count of 1000.\\
       
    \end{minipage}
\end{figure}

\begin{figure}[h!]
    \begin{minipage}[b]{0.25\textwidth}
        \centering
        \includegraphics[width=\linewidth]{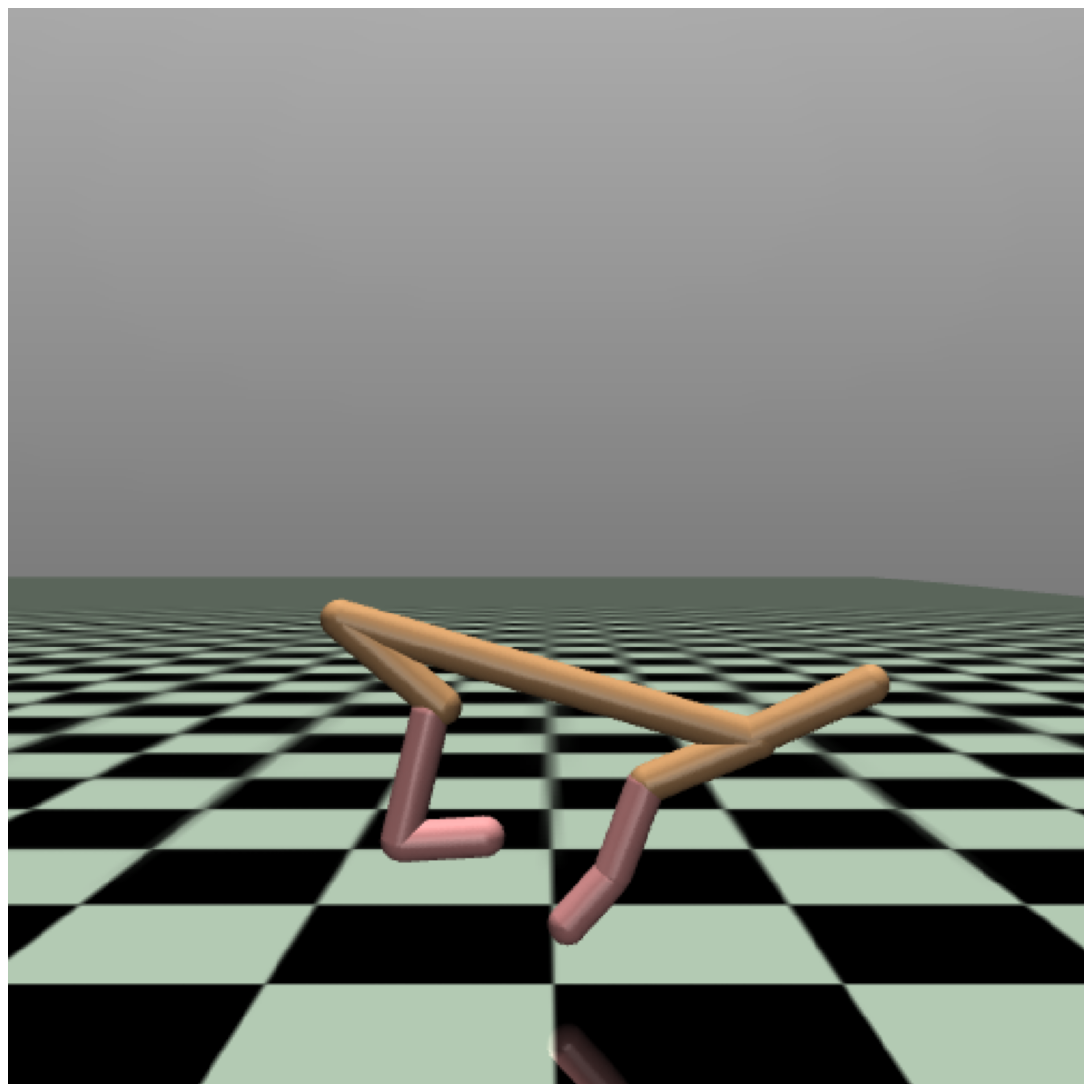}
        \caption{Halfcheetah-v3}
        \label{fig:halfcheetah}
    \end{minipage}%
    \quad
    \begin{minipage}[b]{0.6\textwidth}
    \small
\textbf{\textit{State-action space}}: $\mathcal{S} \in \mathbb{R}^{17}$, $\mathcal{A} \in \mathbb{R}^{6}$.\\

\textbf{\textit{Objective.}} Achieve maximum forward velocity with smooth, coordinated movements.\\

\textbf{\textit{Initialization.}} The agent starts with a slight forward tilt and randomized joint noise.\\

\textbf{\textit{Termination.}} The episode ends after 1000 steps or if the agent's head touches the ground.\\
       
    \end{minipage}
\end{figure}

\begin{figure}[h!]
    \begin{minipage}[b]{0.25\textwidth}
        \centering
        \includegraphics[width=\linewidth]{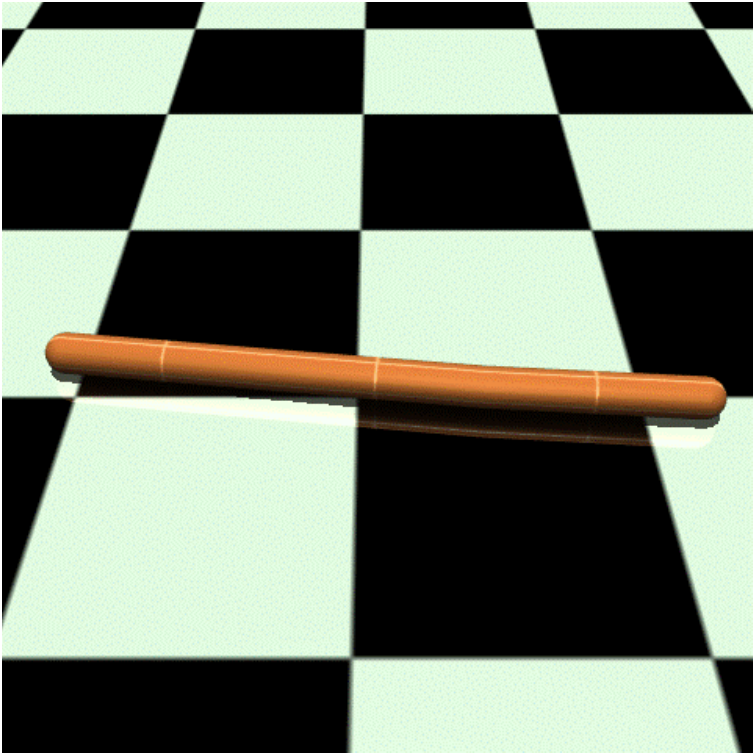}
        \caption{Swimmer-v3}
        \label{fig:swimmer}
    \end{minipage}%
    \quad
    \begin{minipage}[b]{0.6\textwidth}
    \small
\textbf{\textit{State-action space}}: $\mathcal{S} \in \mathbb{R}^{8}$, $\mathcal{A} \in \mathbb{R}^{2}$.\\

\textbf{\textit{Objective.}} Propel forward through water-like dynamics using sinusoidal wave patterns.\\

\textbf{\textit{Initialization.}} The swimmer starts in a straight posture with minor random perturbations.\\

\textbf{\textit{Termination.}} The episode ends after 1000 steps, with no explicit termination for falling.\\
       
    \end{minipage}
\end{figure}

\begin{figure}[h!]
    \begin{minipage}[b]{0.25\textwidth}
        \centering
        \includegraphics[width=\linewidth]{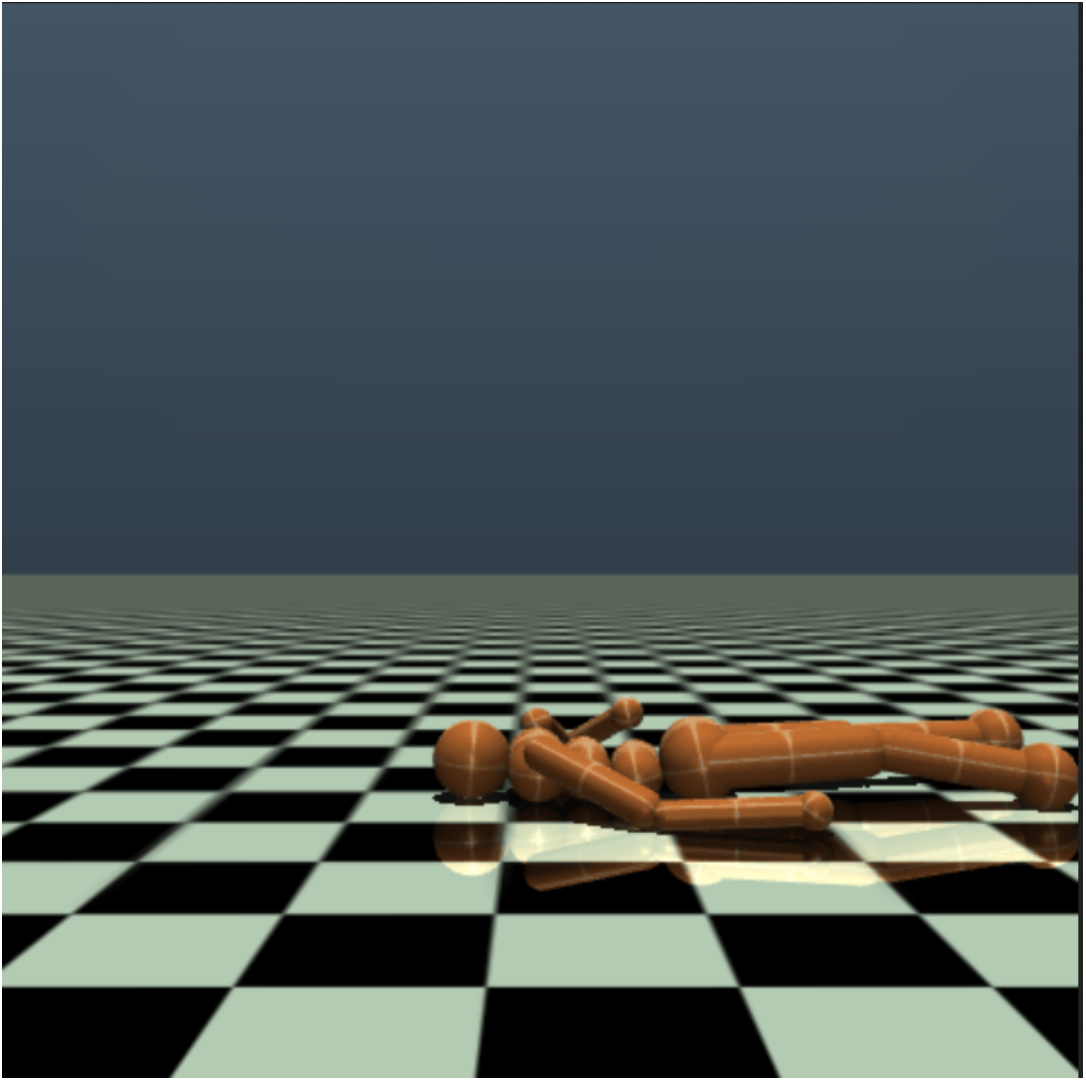}
        \caption{HumanoidStandup}
        \label{fig:humanoidstandup}
    \end{minipage}%
    \quad
    \begin{minipage}[b]{0.6\textwidth}
    \small
\textbf{\textit{State-action space}}: $\mathcal{S} \in \mathbb{R}^{348}$, $\mathcal{A} \in \mathbb{R}^{17}$.\\

\textbf{\textit{Objective.}} Stand up from lying on the ground by applying torques to the joints, with rewards for upward movement and penalties for large actions or strong impacts.\\

\textbf{\textit{Initialization.}} The humanoid starts lying down, with small random noise added to joint positions and velocities.\\ 

\textbf{\textit{Termination.}} The episode does not terminate early; it ends after 1000 steps.\\
       
    \end{minipage}
\end{figure} 

\begin{figure}[h!]
    \begin{minipage}[b]{0.25\textwidth}
        \centering
        \includegraphics[width=\linewidth]{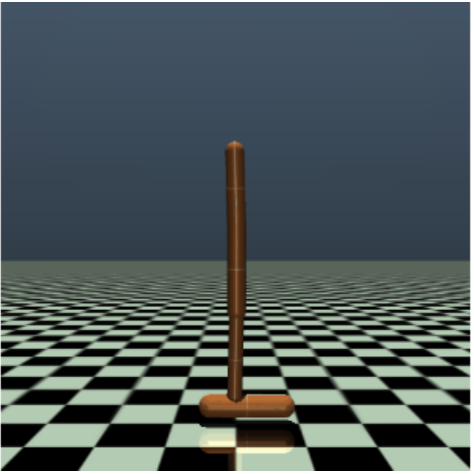}
        \caption{Hopper-v3}
        \label{fig:hopper}
    \end{minipage}%
    \quad
    \begin{minipage}[b]{0.6\textwidth}
    \small
\textbf{\textit{State-action space}}: $\mathcal{S} \in \mathbb{R}^{11}$, $\mathcal{A} \in \mathbb{R}^{3}$.\\

\textbf{\textit{Objective.}} Hop forward as fast as possible by applying torques to the thigh, leg, and foot joints, while staying upright.\\ 

\textbf{\textit{Initialization.}} The hopper starts standing upright with small random perturbations in position and velocity.\\ 

\textbf{\textit{Termination.}} The episode ends if the hopper falls (body hits the ground) or after 1000 steps.\\
    \end{minipage}
\end{figure}

\begin{figure}[h!]
    \begin{minipage}[b]{0.25\textwidth}
        \centering
        \includegraphics[width=\linewidth]{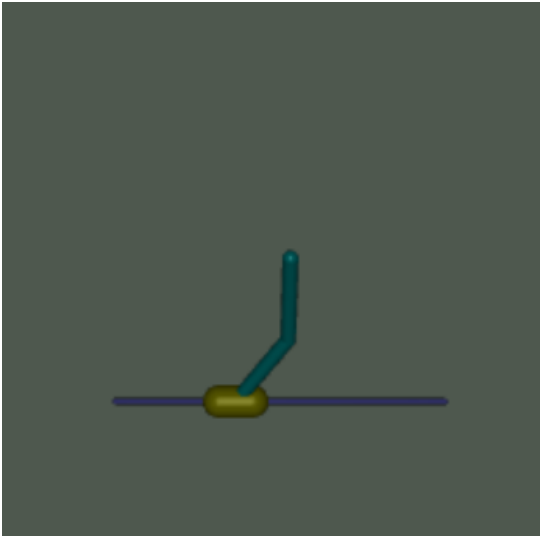}
        \caption{IDP-v3}
        \label{fig:idp}
    \end{minipage}%
    \quad
    \begin{minipage}[b]{0.6\textwidth}
    \small
\textbf{\textit{State-action space}}: $\mathcal{S} \in \mathbb{R}^{9}$, $\mathcal{A} \in \mathbb{R}^{1}$.\\

\textbf{\textit{Objective.}} Balance the second pole upright by applying horizontal forces to the cart, while maximizing time alive and minimizing tip distance and joint velocities.\\ 

\textbf{\textit{Initialization.}} The cart and poles start near the upright position with small random noise in position and velocity.\\ 

\textbf{\textit{Termination.}} The episode ends if the tip of the second pole falls below height 1. Otherwise, it is truncated after 1000 steps.\\
       
    \end{minipage}
\end{figure}

\section{Experimental Hyperparameters}
\vspace{2ex}
\label{app:hyper}
The hyperparameters of all baseline algorithms except the diffusion-based algorithm are shown in Table \ref{tab:baseline hyper}. Additionally, the parameters for all diffusion-based algorithms, including DACERv2, are presented in Table \ref{tab:DACERv2 hyper} and Table \ref{tab:diffusion hyper}. 

The hyperparameter $c, d$ for time-weighted mechanism is determined by the diffusion step size, inspired by the variance-preserving beta schedule used in DDPM \citep{ho2020denoising}. The code of implementation is as follows:
\begin{verbatim}
def vp_alpha_schedule(timesteps: int, b_min=0.1, b_max=10.):
    T = timesteps
    t = np.arange(1, T + 1)
    return np.exp(-b_min / T - 0.5 * (b_max - b_min) * (2 * t - 1) / T ** 2)

# Set parameters
timesteps = 5
alphas = vp_alpha_schedule(timesteps)

# Reverse the alpha array as in B.alphas[self.agent.num_timesteps - 1 - t]
reversed_alphas = alphas[::-1]
t_vals = np.arange(timesteps)

# Fit the exponential form exp(ct + d)
params, _ = curve_fit(exp_fit, t_vals, reversed_alphas)
c, d = params
\end{verbatim}

\begin{table}[ht!]
\centering
\captionsetup{justification=centering,labelsep=newline,font={small,sc}}
\caption{Baseline hyperparameters.}
\label{tab:baseline hyper}
\begin{tabular}{lc}
\toprule
Hyperparameters & Value \\
\hline
\emph{Shared} & \\
\quad Replay buffer capacity & 1,000,000 \\
\quad Buffer warm-up size & 30,000 \\
\quad Batch size & 256 \\
\quad Action bound & $[-1, 1]$ \\
\quad Hidden layers in critic network & [256, 256, 256] \\
\quad Hidden layers in actor network & [256, 256, 256] \\
\quad Activation in critic network & GeLU \\
\quad Activation in actor network & GeLU \\
\quad Optimizer &  Adam ($\beta_{1}=0.9, \beta_{2}=0.999$)\\
\quad Actor learning rate & $1{\rm{e-}}4 $\\
\quad Critic learning rate & $1{\rm{e-}}4 $\\
\quad Discount factor ($\gamma$) & 0.99\\
\quad Policy update interval & 2\\
\quad Target smoothing coefficient ($\rho$) & 0.005\\
\quad Reward scale & 0.2\\
\hline
\emph{Maximum-entropy framework} &\\ 
\quad  Learning rate of $\alpha$ &  $3{\rm{e-}}4 $ \\
\quad  Expected entropy ($\overline{\mathcal{H}}$) &  $\overline{\mathcal{H}}=-{\rm{dim}}(\mathcal{A})$ \\
\hline
\emph{Deterministic policy} &\\ 
\quad Exploration noise&  $\epsilon \sim \mathcal{N}(0,0.1^2)$\\
\hline
\emph{Off-policy} &\\ 
\quad Replay buffer size & $1\times10^6$\\
\quad Sample batch size &  20 \\
\hline
\emph{On-policy} &\\ 
\quad Sample batch size &  2,000 \\
\quad Replay batch size &  2,000 \\
\bottomrule
\end{tabular}
\end{table}

\begin{table}[ht!]
\centering
\captionsetup{justification=centering,labelsep=newline,font={small,sc}}
\caption{Hyperparameter $\eta$ used in DACERv2.}
\label{tab:DACERv2 hyper}
\begin{tabular}{@{}lccccccccc@{}}
\toprule
\textbf{Task} & Hopper & Ant & HalfCheetah & Walker2d & MultiGoal & Hum.\ S. & Humanoid & Swimmer & IDP \\
\midrule
$\eta$ & 1.0 & 1.0 & 1.0 & 1.0 & 1.0 & 0.01 & 0.01 & 0.01 & 0.01 \\
\bottomrule
\end{tabular}
\end{table}

\begin{table}[ht!]
\centering
\captionsetup{justification=centering,labelsep=newline,font={small,sc}}
\caption{Diffusion-based algorithms' hyperparameters}
\label{tab:diffusion hyper}
\begin{tabular}{@{}lcccccc@{}}
\toprule
\textbf{Parameter} & \textbf{DACERv2} & \textbf{DACER} & \textbf{QVPO} & \textbf{QSM}& \textbf{DIME}  & \textbf{DIPO}\\ 
\midrule
Replay buffer capacity & 1e6 & 1e6 & 1e6 & 1e6 &1e6&1e6\\
Buffer warm-up size & 3e4 & 3e4 & 3e4 & 3e4 &3e4&3e4\\
Batch size & 256 & 256 & 256 & 256 & 256 &256 \\
Discount $\gamma$ & 0.99 & 0.99 & 0.99 & 0.99 & 0.99&0.99 \\
Target network soft-update rate $\rho$ & 0.005 & 0.005 & 0.005 & 0.005 & N/A & 0.005\\
Network update times per iteration & 1 & 1&1&1&1&1 \\
Action bound & $[-1, 1]$ & $[-1, 1]$ & $[-1, 1]$ & $[-1, 1]$ & $[-1, 1]$ & $[-1, 1]$\\
Reward scale & 0.2 & 0.2 & 0.2 & 0.2 & 0.2&0.2\\
No. of Actor layers &2 &2 &2 &2 &2&2\\
No. of Actor hidden dims &256 &256 &256 &256&256&256 \\
No. of Critic layers &2 &2 &2 &2 &2&2\\
No. of Critic hidden dims &256 &256 &256 &256&2048&256 \\
Activations in critic network & GeLU & GeLU & Mish & ReLU & ReLU& Mish\\
Activations in actor network & Mish & Mish & Mish & ReLU &ReLU & Mish \\
\textbf{Diffusion steps} & \textbf{5} & 20 & 20 & 20 & 16 & 20\\
Policy delay update & 2 & 2&2&2 &2& 2\\
Action gradient steps & N/A & N/A & N/A & N/A &N/A& 20 \\
No. of Gaussian distributions & 3 & 3 & N/A & N/A &N/A& N/A\\
No. of action samples & 200& 200 & N/A & N/A &N/A&N/A\\
Time-weighted hyperparameter $c$ & 0.4 &N/A & N/A & N/A&N/A&N/A\\
Time-weighted hyperparameter $d$ & -1.8 &N/A & N/A & N/A&N/A&N/A\\
Alpha delay update & 10,000 & 10,000 & N/A & N/A&N/A&N/A\\ 
Noise scale $\lambda$ & 0.1& 0.1 & N/A & N/A &N/A&N/A\\
Optimizer & Adam & Adam & Adam & Adam &Adam&Adam\\
Actor learning rate & $1 \cdot 10^{-4}$ & $1 \cdot 10^{-4}$ & $1 \cdot 10^{-4}$ & $1 \cdot 10^{-4}$ &$3 \cdot 10^{-4}$& $1 \cdot 10^{-4}$\\
Critic learning rate & $1 \cdot 10^{-4}$ & $1 \cdot 10^{-4}$ & $1 \cdot 10^{-4}$ & $1 \cdot 10^{-4}$ &$3 \cdot 10^{-4}$& $1 \cdot 10^{-4}$\\
Alpha learning rate & $3 \cdot 10^{-2}$ & $3 \cdot 10^{-2}$ & N/A & N/A &$1 \cdot 10^{-3}$& N/A\\
Target entropy & $-{\rm{dim}}(\mathcal{A})$ & $ -{\rm{dim}}(\mathcal{A})$ & N/A & N/A & $-4{\rm{dim}}(\mathcal{A})$ &N/A\\
\bottomrule
\end{tabular}
\end{table}

\section{Limitation and Future Work}
\label{sec:limitation}
In this study, we propose the Q-gradient field objective as an auxiliary training loss to provide more informative gradient signals for guiding the diffusion policy. However, algorithms such as PPO \citep{schulman2017proximal} and GRPO \citep{shao2024deepseekmath} do not explicitly learn a Q-function, making it challenging to directly integrate the diffusion policy of DACERv2 and its associated loss function with these methods. This indicates that the generality of our method is currently affected by the presence of value functions. Future work could explore reformulating the auxiliary objective into a purely trajectory-based form, thereby enabling integration with methods that rely solely on policy gradients.

\section{Extra ablation study}
\label{app: extra ablation study}
\begin{figure}[ht!]
\centering
    \includegraphics[width=0.5\textwidth]{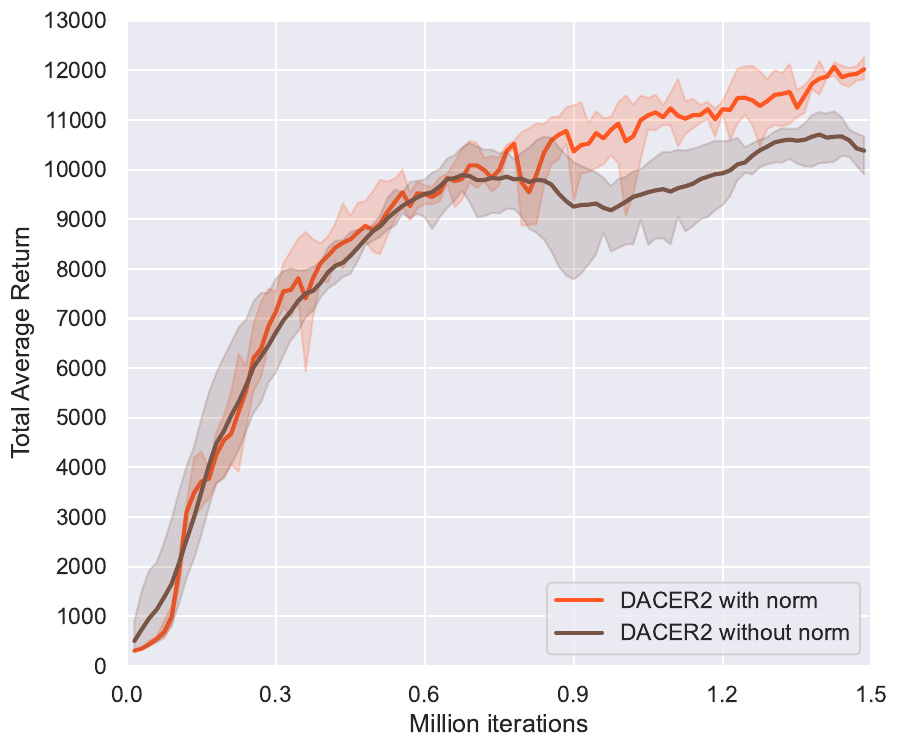} 
    \caption{\centering{Ablation on the normalization of Q-function.}}
\label{fig: norm_TAR}
\end{figure}

We conducted an ablation study on the Humanoid-v3 task to examine the effect of normalizing the Q-gradient. The results presented in Fig.~\ref{fig: norm_TAR} demonstrate that normalization method consistently enhance performance returns.

\newpage
\section{LLM statement}
Large Language Models (LLMs) were employed solely for language refinement in this paper. 
Specifically, we used them to polish grammar, improve clarity, and enhance the academic style of our writing. The role of LLMs was limited to editing and improving the presentation of the text, without contributing to the technical content.

\end{document}

%% file: math_commands.tex
\usepackage{amsmath,amsfonts,bm}

\def\eqref#1{equation~\ref{#1}}

\def\1{\bm{1}}

\DeclareMathAlphabet{\mathsfit}{\encodingdefault}{\sfdefault}{m}{sl}
\SetMathAlphabet{\mathsfit}{bold}{\encodingdefault}{\sfdefault}{bx}{n}













%% file: iclr2026_conference.bbl
\begin{thebibliography}{36}
\providecommand{\natexlab}[1]{#1}
\providecommand{\url}[1]{\texttt{#1}}
\expandafter\ifx\csname urlstyle\endcsname\relax
  \providecommand{\doi}[1]{doi: #1}\else
  \providecommand{\doi}{doi: \begingroup \urlstyle{rm}\Url}\fi

\bibitem[Ada et~al.(2024)Ada, Oztop, and Ugur]{ada2024diffusion}
Suzan~Ece Ada, Erhan Oztop, and Emre Ugur.
\newblock Diffusion policies for out-of-distribution generalization in offline
  reinforcement learning.
\newblock \emph{IEEE Robotics and Automation Letters}, 2024.

\bibitem[Brockman(2016)]{brockman2016openai}
G~Brockman.
\newblock Openai gym.
\newblock \emph{arXiv preprint arXiv:1606.01540}, 2016.

\bibitem[Brockman et~al.(2016)Brockman, Cheung, Pettersson, Schneider,
  Schulman, Tang, and Zaremba]{2012MuJoCo}
Greg Brockman, Vicki Cheung, Ludwig Pettersson, Jonas Schneider, John Schulman,
  Jie Tang, and Wojciech Zaremba.
\newblock Openai gym.
\newblock \emph{arXiv preprint arXiv:1606.01540}, 2016.

\bibitem[Celik et~al.(2025)Celik, Li, Blessing, Li, Palenicek, Peters,
  Chalvatzaki, and Neumann]{celik2025dime}
Onur Celik, Zechu Li, Denis Blessing, Ge~Li, Daniel Palenicek, Jan Peters,
  Georgia Chalvatzaki, and Gerhard Neumann.
\newblock Dime: Diffusion-based maximum entropy reinforcement learning.
\newblock \emph{arXiv preprint arXiv:2502.02316}, 2025.

\bibitem[Chen et~al.(2024)Chen, Liu, Xie, and He]{chen2024deconstructing}
Xinlei Chen, Zhuang Liu, Saining Xie, and Kaiming He.
\newblock Deconstructing denoising diffusion models for self-supervised
  learning.
\newblock \emph{arXiv preprint arXiv:2401.14404}, 2024.

\bibitem[Chen et~al.(2023)Chen, Li, and Zhao]{chen2023boosting}
Yuhui Chen, Haoran Li, and Dongbin Zhao.
\newblock Boosting continuous control with consistency policy.
\newblock \emph{arXiv preprint arXiv:2310.06343}, 2023.

\bibitem[Ding et~al.(2024)Ding, Hu, Zhang, Ren, Zhang, Yu, Wang, and
  Shi]{ding2024diffusion}
Shutong Ding, Ke~Hu, Zhenhao Zhang, Kan Ren, Weinan Zhang, Jingyi Yu, Jingya
  Wang, and Ye~Shi.
\newblock Diffusion-based reinforcement learning via q-weighted variational
  policy optimization.
\newblock \emph{Advances in Neural Information Processing Systems}, 2024.

\bibitem[Duan et~al.(2021)Duan, Guan, Li, Ren, Sun, and
  Cheng]{duan2021distributional}
Jingliang Duan, Yang Guan, Shengbo~Eben Li, Yangang Ren, Qi~Sun, and Bo~Cheng.
\newblock Distributional soft actor-critic: Off-policy reinforcement learning
  for addressing value estimation errors.
\newblock \emph{IEEE Transactions on Neural Networks and Learning Systems},
  33\penalty0 (11):\penalty0 6584--6598, 2021.

\bibitem[Duan et~al.(2025)Duan, Wang, Xiao, Gao, Li, Liu, Zhang, Cheng, and
  Li]{duan2023dsac}
Jingliang Duan, Wenxuan Wang, Liming Xiao, Jiaxin Gao, Shengbo~Eben Li, Chang
  Liu, Ya-Qin Zhang, Bo~Cheng, and Keqiang Li.
\newblock Distributional soft actor-critic with three refinements.
\newblock \emph{IEEE Transactions on Pattern Analysis and Machine
  Intelligence}, 47\penalty0 (5):\penalty0 3935--3946, 2025.
\newblock \doi{10.1109/TPAMI.2025.3537087}.

\bibitem[Fujimoto et~al.(2018)Fujimoto, Hoof, and
  Meger]{fujimoto2018addressing}
Scott Fujimoto, Herke Hoof, and David Meger.
\newblock Addressing function approximation error in actor-critic methods.
\newblock In \emph{International Conference on Machine Learning}, pp.\
  1587--1596. PMLR, 2018.

\bibitem[Fujimoto et~al.(2019)Fujimoto, Meger, and Precup]{fujimoto2019off}
Scott Fujimoto, David Meger, and Doina Precup.
\newblock Off-policy deep reinforcement learning without exploration.
\newblock In \emph{International Conference on Machine Learning}, pp.\
  2052--2062. PMLR, 2019.

\bibitem[Haarnoja et~al.(2017)Haarnoja, Tang, Abbeel, and
  Levine]{haarnoja2017reinforcement}
Tuomas Haarnoja, Haoran Tang, Pieter Abbeel, and Sergey Levine.
\newblock Reinforcement learning with deep energy-based policies.
\newblock In \emph{International Conference on Machine Learning}, pp.\
  1352--1361. PMLR, 2017.

\bibitem[Haarnoja et~al.(2018)Haarnoja, Zhou, Abbeel, and
  Levine]{haarnoja2018soft}
Tuomas Haarnoja, Aurick Zhou, Pieter Abbeel, and Sergey Levine.
\newblock Soft actor-critic: Off-policy maximum entropy deep reinforcement
  learning with a stochastic actor.
\newblock In \emph{International Conference on Machine Learning}, pp.\
  1861--1870. PMLR, 2018.

\bibitem[Hinton(2002)]{hinton2002training}
Geoffrey~E Hinton.
\newblock Training products of experts by minimizing contrastive divergence.
\newblock \emph{Neural computation}, 14\penalty0 (8):\penalty0 1771--1800,
  2002.

\bibitem[Ho et~al.(2020)Ho, Jain, and Abbeel]{ho2020denoising}
Jonathan Ho, Ajay Jain, and Pieter Abbeel.
\newblock Denoising diffusion probabilistic models.
\newblock \emph{Advances in Neural Information Processing Systems},
  33:\penalty0 6840--6851, 2020.

\bibitem[Kang et~al.(2023)Kang, Ma, Du, Pang, and Yan]{kang2023efficient}
Bingyi Kang, Xiao Ma, Chao Du, Tianyu Pang, and Shuicheng Yan.
\newblock Efficient diffusion policies for offline reinforcement learning.
\newblock \emph{Advances in Neural Information Processing Systems}, 36, 2023.

\bibitem[Kumar et~al.(2020)Kumar, Zhou, Tucker, and
  Levine]{kumar2020conservative}
Aviral Kumar, Aurick Zhou, George Tucker, and Sergey Levine.
\newblock Conservative q-learning for offline reinforcement learning.
\newblock \emph{Advances in Neural Information Processing Systems},
  33:\penalty0 1179--1191, 2020.

\bibitem[Li(2023)]{li2023reinforcement}
S~Eben Li.
\newblock \emph{Reinforcement Learning for Sequential Decision and Optimal
  Control}.
\newblock Springer Verlag, Singapore, 2023.

\bibitem[Li et~al.(2024)Li, Krohn, Chen, Ajay, Agrawal, and
  Chalvatzaki]{li2024learning}
Steven Li, Rickmer Krohn, Tao Chen, Anurag Ajay, Pulkit Agrawal, and Georgia
  Chalvatzaki.
\newblock Learning multimodal behaviors from scratch with diffusion policy
  gradient.
\newblock \emph{Advances in Neural Information Processing Systems},
  37:\penalty0 38456--38479, 2024.

\bibitem[Lu et~al.(2022)Lu, Zhou, Bao, Chen, Li, and Zhu]{lu2022dpm}
Cheng Lu, Yuhao Zhou, Fan Bao, Jianfei Chen, Chongxuan Li, and Jun Zhu.
\newblock Dpm-solver: A fast ode solver for diffusion probabilistic model
  sampling in around 10 steps.
\newblock \emph{Advances in Neural Information Processing Systems},
  35:\penalty0 5775--5787, 2022.

\bibitem[Lu et~al.(2025)Lu, Han, Shen, and Li]{lu2025makes}
Haofei Lu, Dongqi Han, Yifei Shen, and Dongsheng Li.
\newblock What makes a good diffusion planner for decision making?
\newblock In \emph{The Thirteenth International Conference on Learning
  Representations}, 2025.

\bibitem[Ma et~al.(2025)Ma, Chen, Wang, Li, and Dai]{ma2025soft}
Haitong Ma, Tianyi Chen, Kai Wang, Na~Li, and Bo~Dai.
\newblock Soft diffusion actor-critic: Efficient online reinforcement learning
  for diffusion policy.
\newblock \emph{arXiv preprint arXiv:2502.00361}, 2025.

\bibitem[Messaoud et~al.(2024)Messaoud, Mokeddem, Xue, Pang, An, Chen, and
  Chawla]{messaoud2024s}
Safa Messaoud, Billel Mokeddem, Zhenghai Xue, Linsey Pang, Bo~An, Haipeng Chen,
  and Sanjay Chawla.
\newblock S2ac: Energy-based reinforcement learning with stein soft actor
  critic.
\newblock \emph{arXiv preprint arXiv:2405.00987}, 2024.

\bibitem[Psenka et~al.(2023)Psenka, Escontrela, Abbeel, and
  Ma]{psenka2023learning}
Michael Psenka, Alejandro Escontrela, Pieter Abbeel, and Yi~Ma.
\newblock Learning a diffusion model policy from rewards via q-score matching.
\newblock \emph{arXiv preprint arXiv:2312.11752}, 2023.

\bibitem[Ren et~al.(2024)Ren, Lidard, Ankile, Simeonov, Agrawal, Majumdar,
  Burchfiel, Dai, and Simchowitz]{ren2024DPPO}
Allen~Z Ren, Justin Lidard, Lars~L Ankile, Anthony Simeonov, Pulkit Agrawal,
  Anirudha Majumdar, Benjamin Burchfiel, Hongkai Dai, and Max Simchowitz.
\newblock Diffusion policy policy optimization.
\newblock \emph{arXiv preprint arXiv:2409.00588}, 2024.

\bibitem[Schulman et~al.(2017)Schulman, Wolski, Dhariwal, Radford, and
  Klimov]{schulman2017proximal}
John Schulman, Filip Wolski, Prafulla Dhariwal, Alec Radford, and Oleg Klimov.
\newblock Proximal policy optimization algorithms.
\newblock \emph{arXiv preprint arXiv:1707.06347}, 2017.

\bibitem[Shao et~al.(2024)Shao, Wang, Zhu, Xu, Song, Bi, Zhang, Zhang, Li, Wu,
  et~al.]{shao2024deepseekmath}
Zhihong Shao, Peiyi Wang, Qihao Zhu, Runxin Xu, Junxiao Song, Xiao Bi, Haowei
  Zhang, Mingchuan Zhang, YK~Li, Yang Wu, et~al.
\newblock Deepseekmath: Pushing the limits of mathematical reasoning in open
  language models.
\newblock \emph{arXiv preprint arXiv:2402.03300}, 2024.

\bibitem[Song et~al.(2020)Song, Sohl-Dickstein, Kingma, Kumar, Ermon, and
  Poole]{song2020score}
Yang Song, Jascha Sohl-Dickstein, Diederik~P Kingma, Abhishek Kumar, Stefano
  Ermon, and Ben Poole.
\newblock Score-based generative modeling through stochastic differential
  equations.
\newblock \emph{arXiv preprint arXiv:2011.13456}, 2020.

\bibitem[Sutton \& Barto(2018)Sutton and Barto]{sutton2018reinforcement}
Richard~S Sutton and Andrew~G Barto.
\newblock \emph{Reinforcement learning: An introduction}.
\newblock MIT press, 2018.

\bibitem[Teh et~al.(2016)Teh, Thiery, and Vollmer]{teh2016consistency}
Yee~Whye Teh, Alexandre~H Thiery, and Sebastian~J Vollmer.
\newblock Consistency and fluctuations for stochastic gradient langevin
  dynamics.
\newblock \emph{The Journal of Machine Learning Research}, 17\penalty0
  (1):\penalty0 193--225, 2016.

\bibitem[Van~Hasselt et~al.(2016)Van~Hasselt, Guez, and Silver]{van2016deep}
Hado Van~Hasselt, Arthur Guez, and David Silver.
\newblock Deep reinforcement learning with double q-learning.
\newblock In \emph{Proceedings of the AAAI conference on artificial
  intelligence}, volume~30, 2016.

\bibitem[Wang et~al.(2024)Wang, Wang, Jiang, Zou, Liu, Song, Wang, Xiao, Wu,
  Duan, and Li]{wang2024diffusion}
Yinuo Wang, Likun Wang, Yuxuan Jiang, Wenjun Zou, Tong Liu, Xujie Song, Wenxuan
  Wang, Liming Xiao, Jiang Wu, Jingliang Duan, and Shengbo Li.
\newblock Diffusion actor-critic with entropy regulator.
\newblock \emph{Advances in Neural Information Processing Systems}, 2024.

\bibitem[Wang et~al.(2023)Wang, Hunt, and Zhou]{wang2022diffusion}
Zhendong Wang, Jonathan~J Hunt, and Mingyuan Zhou.
\newblock Diffusion policies as an expressive policy class for offline
  reinforcement learning.
\newblock \emph{The Eleventh International Conference on Learning
  Representations}, 2023.

\bibitem[Welling \& Teh(2011)Welling and Teh]{welling2011bayesian}
Max Welling and Yee~W Teh.
\newblock Bayesian learning via stochastic gradient langevin dynamics.
\newblock In \emph{Proceedings of the 28th international conference on machine
  learning (ICML-11)}, pp.\  681--688. Citeseer, 2011.

\bibitem[Yang et~al.(2023{\natexlab{a}})Yang, Huang, Lei, Zhong, Yang, Fang,
  Wen, Zhou, and Lin]{yang2023policy}
Long Yang, Zhixiong Huang, Fenghao Lei, Yucun Zhong, Yiming Yang, Cong Fang,
  Shiting Wen, Binbin Zhou, and Zhouchen Lin.
\newblock Policy representation via diffusion probability model for
  reinforcement learning.
\newblock \emph{arXiv preprint arXiv:2305.13122}, 2023{\natexlab{a}}.

\bibitem[Yang et~al.(2023{\natexlab{b}})Yang, Srivastava, and
  Mandt]{yang2023diffusion}
Ruihan Yang, Prakhar Srivastava, and Stephan Mandt.
\newblock Diffusion probabilistic modeling for video generation.
\newblock \emph{Entropy}, 25\penalty0 (10):\penalty0 1469, 2023{\natexlab{b}}.

\end{thebibliography}
